\documentclass[11pt]{article}

\usepackage[final]{acl}

\usepackage{times}
\usepackage{latexsym}
\usepackage{amsmath}
\usepackage{booktabs}
\usepackage{multirow}
\usepackage{arydshln}
\usepackage{amssymb} 
\usepackage{adjustbox}
\usepackage{tabularx}
\usepackage[table]{xcolor}
\usepackage{booktabs}
\usepackage{multirow}
\usepackage{graphicx}
\usepackage{arydshln}
 \usepackage[most]{tcolorbox}

\definecolor{oursblue}{RGB}{232,242,255}
\definecolor{bestyellow}{RGB}{255,246,210}

\newcommand{\oursrow}{\rowcolor{oursblue}}
\newcommand{\best}[1]{\cellcolor{bestyellow}\textbf{#1}}

\newcommand{\ph}[1]{\textcolor{red}{\textbf{\{#1\}}}}

\newtcblisting{promptbox}[1][]{
breakable, listing only,
colback=gray!5, colframe=gray!45!black,
fonttitle=\bfseries, coltitle=white,
title={#1},
listing options={
  breaklines=true, breakindent=0pt, breakatwhitespace=false,
  basicstyle=\ttfamily\footnotesize, columns=fullflexible,
  keepspaces=true, showstringspaces=false,
  extendedchars=true
}
}

\newtcblisting{agenticpromptbox}[1][]{
breakable, listing only,
colback=oursblue!60, colframe=blue!45!black,
fonttitle=\bfseries, coltitle=white,
title={#1},
listing options={
  breaklines=true, breakindent=0pt, breakatwhitespace=false,
  basicstyle=\ttfamily\footnotesize, columns=fullflexible,
  keepspaces=true, showstringspaces=false,
  extendedchars=true,
  escapeinside={(*}{*)}
}
}

\newtcblisting{evalpromptbox}[1][]{
breakable, listing only,
colback=bestyellow!80, colframe=orange!50!black,
fonttitle=\bfseries, coltitle=white,
title={#1},
listing options={
  breaklines=true, breakindent=0pt, breakatwhitespace=false,
  basicstyle=\ttfamily\footnotesize, columns=fullflexible,
  keepspaces=true, showstringspaces=false,
  extendedchars=true,
  escapeinside={(*}{*)}
}
}

\usepackage[T1]{fontenc}
\usepackage[utf8]{inputenc}

\usepackage{microtype}

\usepackage{inconsolata}

\usepackage{graphicx}

\title{CAST: Critique-Aware Supervision for Training Reliable Long-Horizon Tool-Calling Agents}
\author{Amir Saeidi$^{\dagger*}$ \quad Zehua Zhang$^\dagger$\thanks{Equal contribution. Correspondence: \href{mailto:ssaeidi1@asu.edu}{\textbf{ssaeidi1@asu.edu}}} \quad Rishitosh Singh$^\dagger$  \quad Naman Ahuja$^\dagger$ \\
\textbf{Vivek Gupta}$^\dagger$ \quad \textbf{Ali Payani}$^\ddagger$ \quad \textbf{Gaowen Liu}$^\ddagger$ \quad  \textbf{Jayanth Srinivasa}$^\ddagger$ \quad \textbf{Chitta Baral}$^\dagger$ \\\\ 
$^\dagger$Arizona State University \quad $^\ddagger$Cisco Research 
}
\begin{document}
\maketitle

\begin{abstract}

Large language model (LLM) agents are increasingly deployed in long-horizon, interactive, and stateful environments. In these settings, a single wrong action, such as refunding the wrong purchase, can cause irreversible task failure and must be intercepted before execution. Such failures may not appear in every single run, but can emerge across repeated trials, making reliability across steps and trials critical. However, ensuring agentic reliability is challenging: even frontier LLMs struggle to explain why an action may be wrong, especially in long, intertwined trajectories governed by domain-specific policies.
Much recent work relies on prompt-based critique agents, while optimization-based methods lack a systematic way to produce rich verification rationales for training. We address this gap with CAST, a critique-aware training framework that converts sparse task outcomes into action-level supervision for critique learning and policy optimization. CAST analyzes agent trajectories to synthesize structured rationales explaining action validity under partial observability. The resulting critique model is used to construct critique-aware training data for optimizing the policy model. Fine-tuning Qwen3-family models on dynamic tool-calling benchmarks, CAST improves reliability across domains, outperforming GPT-OSS-120B by over 10\% pass\textasciicircum{}4 on Retail tasks and yielding an additional 9\% improvement on Telehealth in an out-of-domain setting. These results demonstrate that critique-aware training improves the robustness of LLM agents in realistic dynamic environments.

% Many recent work build critique agents relying on prompting techniques, while existing optimization-based methods lack a systematic approach for producing rich verification rationales. To address this gap, we introduce CAST, a critique-aware training framework that transforms sparse task outcomes into action-level supervision for critique learning and policy optimization. CAST performs agentic trajectory analysis to synthesize structured verification rationales that explain action validity under partial observability. These rationales supervise a critique model, which is then used to identify verified successful experiences for optimizing the policy model. We fine-tune Qwen3-family models and evaluate them on dynamic tool-calling benchmarks. CAST improves reliability across domains, yielding more than a 10\% gain over GPT-OSS-120B on Retail on the pass\textasciicircum{}4 metric and a further 9\% improvement on Telehealth in an out-of-domain setting. Our results demonstrate that critique-aware training improves the robustness of LLM agents in realistic dynamic environments.

\begin{figure}
    \centering
    \includegraphics[width=1\linewidth]{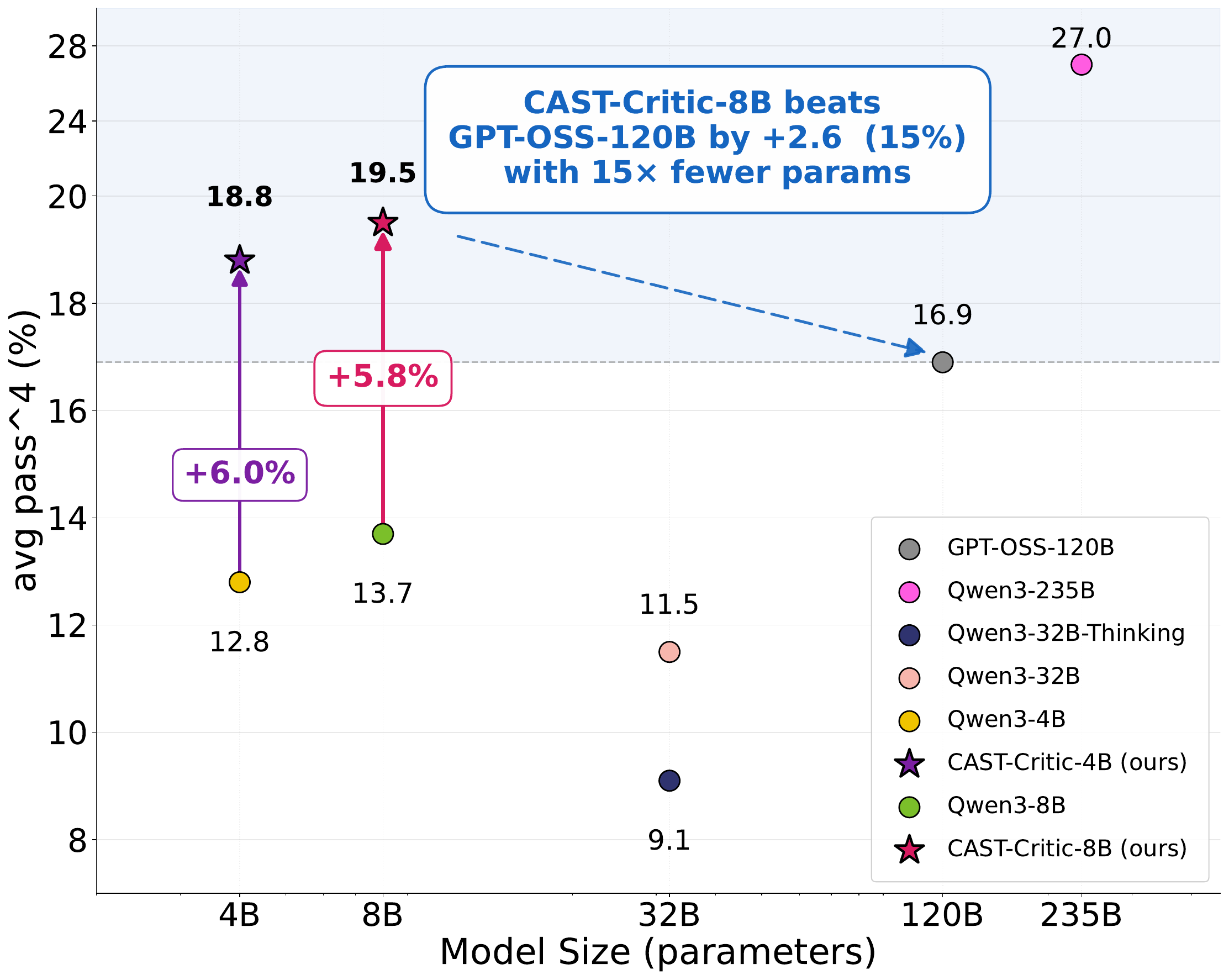}
    \caption{CAST help improve reliability of LLM agents in dynamic environments across trials, which allows smaller models to outperform substantially larger models on reliability-oriented metrics.}
    \label{fig:main_com}
\end{figure}
\end{abstract}

\section{Introduction}
% Large language models excel at reasoning and question answering tasks. The recent shift toward an agentic paradigm has further enabled these models to interact with complex environments, achieving strong performance in tool use and problem solving. Despite these advances, the reliability of language agents remains a critical challenge. Specifically, agents struggle with execution consistency and robustness, frequently exhibiting unstable behavior where they successfully navigate a task in one instance but fail to find a solution in an identical or highly similar setting due to slight variations in the environment or trajectory.

\begin{figure*}[t]
    \centering
    \includegraphics[width=1\linewidth]{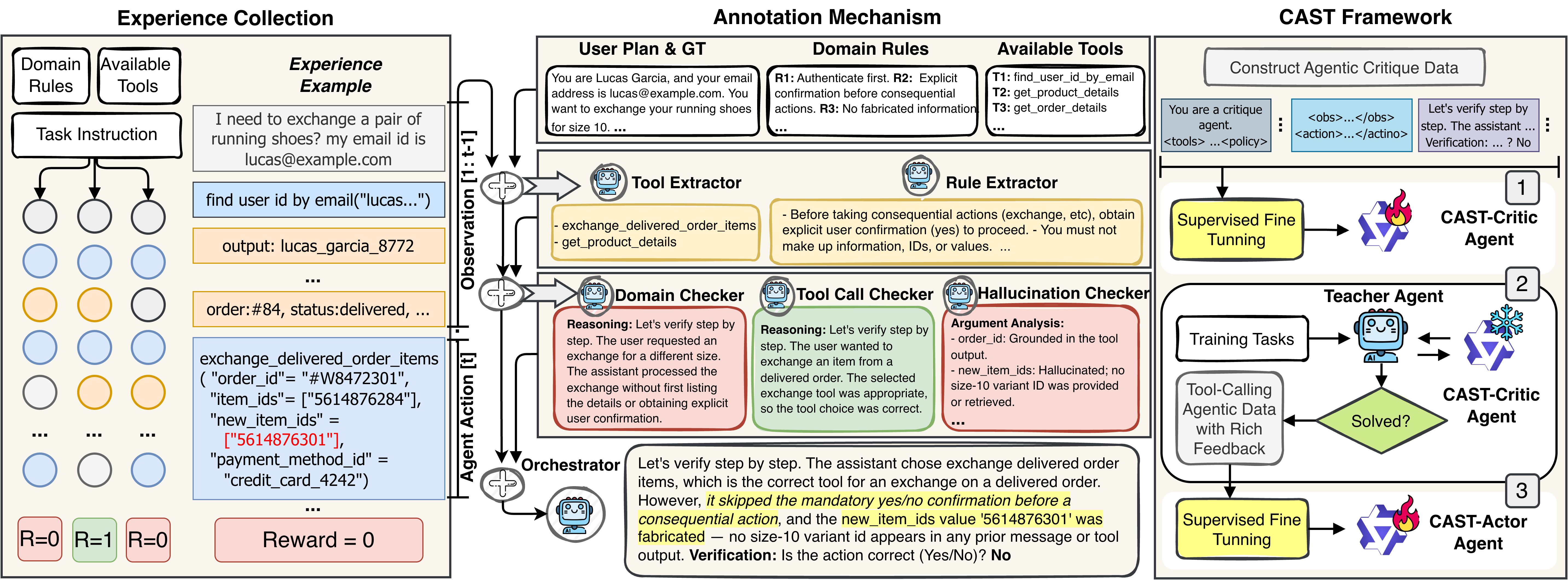}
    \caption{CAST overview pipeline. \textbf{Stage 1} collects trajectories by repeatedly running a teacher policy on the same training tasks. \textbf{Stage 2} applies an agentic verification framework to generate rationales that explain whether each action is valid or problematic. \textbf{Stage 3} uses these annotated trajectories to train a critique agent, then redeploys the critique agent to collect verification-enriched experiences for training a critique-aware tool-calling agent.}
    \label{fig:cast_overview}
\end{figure*}

Large language models have demonstrated strong capabilities in reasoning~\cite{guo2025deepseek}, question answering~\cite{yue2025survey}, and instruction following~\cite{zhou2023instruction}. More recently, the agentic paradigm has extended these capabilities to interactive environments, where models use tools, track evolving states, and make decisions over long horizons~\cite{qin2024toolllm, saeidi2026vulcan}. Despite this progress, reliability remains a central challenge for language agents~\cite{yao2024tau, uddin2026recall}. In dynamic environments, agents may solve a task in one trajectory but fail in an identical or highly similar scenario due to small changes in observations, tool outputs, or previous actions~\cite{barres2025tau}. This inconsistency becomes especially problematic in realistic deployments, where each action can affect the subsequent interaction and repeated attempts are often unavailable. The challenge is amplified in longer conversations, where small mistakes can accumulate and cause task failure~\cite{laban2025llms}.

To address this inconsistency, recent works~\cite{mishra2025can, saeidi2026fama, uddin2026ledgeragent,tan2025process, qian2025userrl} have explored both inference-time agentic mechanisms and learning-based optimization methods. Complex agentic systems can improve decision making through planning, reflection, or self-correction, but they often introduce substantial inference latency, token overhead, and limited scalability~\cite{renze2024self}. Although reinforcement learning methods such as GRPO~\cite{shao2024deepseekmath} reduce inference-time overhead compared with these frameworks and can improve overall task success, single shot performance gains do not necessarily constitute reliability across multiple trials. Natural-language actor-critic methods provide a promising direction by using textual critiques as intermediate feedback, but they typically optimize the actor and critic jointly without sufficient supervision for the critic to produce reliable verification rationales~\cite{jiang2026asymmetric, hongNLAC2025}. This limitation is particularly important because even frontier models can struggle to judge whether an actor's action is wrong in long-horizon interactions.

% To overcome these limitations, we propose CAST, a critique-aware training framework for developing reliable tool-calling agents. Rather than solely relying on sparse trajectory-level rewards, CAST generates an explicit supervision signal for action level verification. It first analyzes trajectories collected from a policy agent and uses an agentic data-generation process to construct structured rationales that identify when and why individual actions are valid or problematic. A critique model is trained on these rationales and then reused during experience collection to annotate new interactions with step-level verification feedback. The resulting critique-enriched successful trajectories are used to optimize the policy agent, yielding a model that can benefit from explicit critique during inference or operate independently as a fine-tuned policy agent.

To overcome these limitations, we propose CAST, a critique-aware training framework for improving long-horizon tool-calling reliability. CAST utilizes a multi-agent system to convert sparse trajectory-level outcomes into structured action-level verification supervision. Trained on it, a critique model (CAST-Critic) learns to judge whether each proposed action is valid under the information available at that step. CAST-Critic is then reused to annotate new interactions with fine-grained verification feedback. 
The resulting critique-enriched successful trajectories are then used to optimize the policy agent (CAST-Policy) that can benefit from explicit critique during inference or operate independently as a fine-tuned policy agent.

We evaluate CAST across four dynamic tool-calling domains, including retail, airline~\cite{yao2024tau}, telecom, and telehealth~\cite{he2025impatient}. Using retail as the in-domain environment, fine-tuned Qwen3~\cite{yang2025qwen3} models improve over base instruct models by 22.6\% on pass\textasciicircum{}1 and 12.94\% on pass\textasciicircum{}4. Across out-of-domain settings, CAST achieves average gains of 3.7\% on pass\textasciicircum{}1 and 5.1\% on pass\textasciicircum{}4. In addition, CAST achieves performance comparable to resource-intensive agentic frameworks while substantially reducing token usage and inference latency. These results show that structured critique supervision can improve both the reliability and efficiency of LLM agents in dynamic long-horizon environments. Our main contributions can be summarized as follows:
\paragraph{1.} We propose CAST, a scalable critique-aware training framework that converts trajectory-level outcomes into structured action-level verification signals for training tool-calling agents.
\paragraph{2.} We introduce an agentic data-generation pipeline for producing high-quality step-level critique rationales, which are used to train a critique model and construct critique-enriched policy optimization data.
\paragraph{3.} We conduct extensive evaluations across two models and four domains, showing that CAST improves agent reliability while reducing the cost of inference compared with more resource-intensive agentic baselines.
\section{Related Work}
\label{sec:related_work}
% \paragraph{Tool-Augmented LLM Agents.} 
% The integration of external tools has significantly expanded the capabilities of large language models, transforming them into interactive agents capable of multi-step problem solving \cite{yao2022react, schick2023toolformer}. Despite these architectural advances, multi-turn tool execution remains brittle. In environments requiring long-horizon planning, early reasoning flaws, hallucinations, and behavior drift frequently cascade, leading to unrecoverable task failures \cite{toolsandbox2025, yu2026wildtoolbench}. Consequently, a core challenge in LLM tool-using research is developing mechanisms to accurately monitor, evaluate, and correct intermediate reasoning steps before execution errors compound.

\paragraph{Tool-Augmented LLM Agents.}
Tool-augmented LLMs extend language models into interactive agents that plan and act over multi-step environments \cite{yao2022react, schick2023toolformer}. While single-turn tool use is increasingly reliable, performance degrades sharply in long-horizon interactions, where hallucinated arguments, domain-rule violations, and incorrect tool calls compound across turns into unrecoverable failures \cite{lu2025toolsandbox, yu2026wildtoolbench}. Recent reliability-oriented benchmarks such as $\tau$-Bench \cite{yao2024tau} expose this gap directly by measuring whether agents succeed \emph{consistently} across repeated runs of the same task, which is non-trivial to certain tasks that requires reliability. Closing this gap requires mechanisms that verify intermediate actions \emph{before} errors propagate, rather than scoring only the final trajectory outcome.

\paragraph{Language-Based Feedback and Self-Correction.}
While task rewards are often sparse and binary, recent work supervises LLM behavior with natural-language feedback \cite{lin2024criticbench}. Frameworks such as Reflexion and Self-Refine let models iteratively revise their outputs from textual critiques generated after each attempt \cite{shinn2023reflexion, madaan2023selfrefine}. Two limitations make these methods difficult to deploy in long-horizon tool use. Critiques are produced post-hoc, Reflexion requires a terminal reward and Self-Refine requires a finished output, and neither formalizes the information available to the agent at the moment an action is taken, so when transplanted to multi-turn tool use the critic implicitly conditions on later observations or ground-truth user intent that were not part of the LLM's context \cite{li2025dancing}. Recent methods also improve reliability through failure recovery or modular optimization. PALADIN uses failure-conditioned retrieval, while EvoTool evolves tool-use policies from trajectory-level feedback \citep{vuddanti2025paladin, yang2026evotool}. These methods still provide coarser supervision than direct action-level verification. This motivates a critic that judges each action using only the partial history available at that step.

\paragraph{Actor-Critic Paradigms with Textual Supervision.}
Recent work has attempted to formalize language-based feedback within actor-critic architectures. Natural Language Actor-Critic \citep{hongNLAC2025} trains a generative critic to produce textual feedback rather than value estimates, and the Asymmetric Actor-Critic framework of \citet{jiang2026asymmetric} pairs a fixed proprietary actor with a lightweight open-source
critic for multi-turn reliability. More broadly, generative verifiers cast reward modeling as next-token prediction so each judgment contains a Chain-of-Thought rationale~\citep{zhang2025generative}, and process-supervised RL uses an LLM judge for turn-level credit assignment in long-horizon tool-integrated reasoning~\citep{tan2025process}. These approaches either jointly optimize actor and critic under sparse rewards or rely on heavyweight RL pipelines, leaving the critic without supervision for why an intermediate action is unreliable.

% CAST instead decouples the two: an agentic verification procedure produces structured rationales that train a dedicated critique model, whose outputs then annotate new trajectories for SFT-based policy optimization, retaining action-level, partially observable judgments while avoiding RL instability.

% Complementary approaches improve tool-use reliability through failure-aware recovery and modular policy optimization. 
% PALADIN trains agents on annotated trajectories with injected failures and recovery.
% During inference, it retrieves predefined examples related to the current failure reason to recover from tool errors, which may introduce additional instability and cost~\citep{vuddanti2025paladin}.
% EvoTool uses trajectory-level annotated blame attribution and natural-language feedback to evolve a designated tool-use module while keeping model weights fixed~\citep{yang-etal-2026-evotool}. 
% It requires repeated inference for prompt mutation and population selection, introducing optimization latency and providing coarser supervision than action-level verification.

% CAST instead trains a dedicated critic to verify proposed actions from the partial history, then generate structured critiques to guide inference and train a standalone critique-aware policy.

\section{Method}

In long-horizon tasks, an agent may follow a largely valid trajectory, yet a single erroneous action can jeopardize task completion by triggering cascading and sometimes irreversible consequences. For such tasks, it is important to verify individual actions so that incorrect actions can be prevented early or their downstream effects can be remediated before they lead to failure. However, verification is nontrivial: the critique model must judge a proposed action using only the information available at the current step, without access to future observations.
We thus propose CAST, a critique-aware training framework that learns action-level verification signals and uses them to improve tool-calling policies (Figure~\ref{fig:cast_overview}). We formalize the problem setting and then describe the framework in detail as follows. 

% We propose CAST, a critique-aware training framework for learning action-level verification signals and using them to improve tool-calling policies. At a high level, CAST samples multiple trajectories for each training task, applies an agentic verification procedure to synthesize structured rationales and verification labels for individual actions, trains a critique model on this supervision, and uses the learned critique signal to construct policy optimization data (See Figure~\ref{fig:cast_overview}). The following subsections formalize the problem setting and describe each stage of CAST in detail.

\subsection{Problem Definition}

We consider a dynamic tool-calling environment in which an agent interacts with a user over multiple turns. At step $t$, the agent receives the current observation $o_t$, maintains an interaction history $h_t = \big((o_1,a_1), \ldots, (o_{t-1},a_{t-1})\big).$, and selects an action
\[
a_t \sim \pi_\phi(\cdot \mid h_t, o_t, \mathcal{K}),
\]
where $\mathcal{K}$ denotes the fixed tool set and domain constraints provided in the system prompt. The action $a_t$ may either respond to the user or invoke a tool. Executing $a_t$ produces the next observation $o_{t+1}$ and updates the latent environment state.

A trajectory is defined as
\[
\tau=\{(o_t,a_t)\}_{t=1}^{T},
\]
with a trajectory-level success label $Y(\tau)\in\{0,1\}$. Standard policy optimization uses this outcome to learn a policy that maximizes
\[
\max_\phi \; \mathbb{E}_{\tau\sim\pi_\phi}[Y(\tau)].
\]
% In long-horizon tasks, an agent may follow a largely valid trajectory, yet a single erroneous action can jeopardize task completion by triggering cascading and sometimes irreversible consequences. For such tasks, it is preferable to verify individual actions so that incorrect actions can be prevented early or their downstream effects can be remediated before they lead to failure. We aim to learn an action-level critique signal that explains why a proposed action may be problematic, enabling the assistant to revise its plan before the incorrect action is executed and to solve the same task more reliably across repeated runs.

Because $Y(\tau)$ alone does not reveal which action caused a failure, we additionally learn an action-level signal that judges each $a_t$ before errors propagate.

We define an action verifier $g$ as a function that maps the current interaction context and candidate action to a structured verification output
\[
g(h_t,o_t,a_t,\mathcal{K}) \rightarrow (e_t,y_t),
\]
where $y_t\in\{0,1\}$ indicates whether action $a_t$ is valid under the information available at step $t$, and $e_t$ is a structured rationale justifying the verification decision. 
If the verifier accepts the candidate action, the action is executed and the environment advances; otherwise, the verifier’s decision and rationale are returned to the assistant for revision.
Verifier feedback and rejected candidates are not included in the original agent-environment trajectory, but instead form an auxiliary critique trace used before execution. Verification is conditioned only on the current interaction context and does not use future observations ${o_{t+1},\ldots,o_T}$.

CAST learns a parameterized critique model (CAST-Critic)
\[
C_\theta(h_t,o_t,a_t,\mathcal{K})=(\hat e_t,\hat y_t),
\]
and uses its outputs to construct critique-aware training data for optimizing the policy $\pi_\phi$.

\subsection{Overview of CAST}

CAST consists of three learning stages. First, we collect trajectories from an initial policy $\pi_\phi$ over a task set $\mathcal{D}=\{x_i\}_{i=1}^{M}$. 
% For each task $x_i$, we sample $N$ executions
% \[
% \mathcal{T}_i=\{\tau_i^{(1)},\ldots,\tau_i^{(N)}\},
% \]
% and construct the experience buffer 
% \[
% \mathcal{B}=\bigcup_{i=1}^{M}\mathcal{T}_i.
% \]
For each task $x_i$, we sample $N$ executions
$\mathcal{T}_i=\{\tau_i^{(j)}\}_{j=1}^{N}$ and construct the experience buffer
$\mathcal{B}=\{\tau_i^{(j)}: i\in[M], j\in[N]\}$.
Because the environment is dynamic, repeated executions of the same task produce diverse action sequences and outcomes. The buffer $\mathcal{B}$ contains both successful and failed trajectories.

Second, CAST constructs an annotated critique dataset. For each action $a_t$ in each trajectory $\tau\in\mathcal{B}$, an agentic verification procedure $V$ using a teacher model generates
\[
V(h_t,o_t,a_t,\mathcal{K},\Omega) \rightarrow (e_t,y_t).
\]
where $\Omega$ denotes privileged information (if available) observable only at annotation time, such as ground truth action trajectories.
The student critique model $C_\theta$ and the deployed policy never observe $\Omega$ and condition only on the step-local context $c_t=(h_t,o_t,a_t,\mathcal{K})$. This asymmetry allows $V$ to assign reliable labels that are often undecidable from $c_t$ alone to better facilitate critique learning.
The resulting dataset is
\[
\mathcal{A}_{\text{critique}}
=
\{(h_t,o_t,a_t,\mathcal{K},e_t,y_t)\}.
\]

Third, we train CAST-Critic and use it to collect critique-enriched trajectories. Successful critique-enriched trajectories are then selected to optimize the policy model $\pi_\phi$.

\subsection{Agentic Verification}

The verification procedure assigns a structured rationale $e_t$ and binary label $y_t$ to each action. Given context $c_t=(h_t,o_t,a_t,\mathcal{K})$, the verifier determines whether $a_t$ is supported by the information available at step $t$.

We decompose action failures into three categories:
hallucination $f_{\text{hall}}(c_t)$, domain violation $f_{\text{domain}}(c_t)$, and wrong tool usage $f_{\text{tool}}(c_t)$. Hallucination captures unsupported assumptions, fabricated information, or arguments inconsistent with the observed state. Domain violation captures actions that conflict with system rules, task requirements, or domain-specific constraints. Wrong tool usage captures cases where the agent selects a tool that is incorrect or inconsistent with the current context.
The final verification label is computed as
\[
y_t =
\mathbf{1}
[
\neg f_{\text{hall}}(c_t)
\wedge
\neg f_{\text{domain}}(c_t)
\wedge
\neg f_{\text{tool}}(c_t)
].
\]
The rationale $e_t$ explains the verification decision by grounding it in the detected failure type or confirming that no failure is found.

\begin{table*}[t]
\centering
\scriptsize
\setlength{\tabcolsep}{3pt}
\resizebox{\textwidth}{!}{
% \begin{tabular}{lcccccccccccc}
\begin{tabular}{l|ccc|ccc|ccc|ccc}
% \toprule
% \multirow{2}{*}{\textbf{Model}} 
% & \multicolumn{3}{c}{$\tau$-Retail (in-domain)} 
% & \multicolumn{3}{c}{$\tau$-Airline (out-of-domain)} 
% & \multicolumn{3}{c}{$\tau$-Telecom (out-of-domain)} 
% & \multicolumn{3}{c}{$\tau$-Telehealth (out-of-domain)} \\
% \cmidrule(lr){2-4}
% \cmidrule(lr){5-7}
% \cmidrule(lr){8-10}
% \cmidrule(lr){11-13}
% & Pass\textasciicircum{}1 & Pass\textasciicircum{}3 & Pass\textasciicircum{}4
% & Pass\textasciicircum{}1 & Pass\textasciicircum{}3 & Pass\textasciicircum{}4
% & Pass\textasciicircum{}1 & Pass\textasciicircum{}3 & Pass\textasciicircum{}4
% & Pass\textasciicircum{}1 & Pass\textasciicircum{}3 & Pass\textasciicircum{}4 \\
% \midrule

\toprule
\multirow{2}{*}{\textbf{Model}} 
& \multicolumn{3}{c}{\textbf{$\tau$-Retail (in-domain)}} 
& \multicolumn{3}{c}{\textbf{$\tau$-Airline (out-of-domain)}} 
& \multicolumn{3}{c}{\textbf{$\tau$-Telecom (out-of-domain)}} 
& \multicolumn{3}{c}{\textbf{$\tau$-Telehealth (out-of-domain)}} \\
\cmidrule(lr){2-4}
\cmidrule(lr){5-7}
\cmidrule(lr){8-10}
\cmidrule(lr){11-13}
& \textbf{Pass\textasciicircum{}1} & \textbf{Pass\textasciicircum{}3} & \textbf{Pass\textasciicircum{}4}
& \textbf{Pass\textasciicircum{}1} & \textbf{Pass\textasciicircum{}3} & \textbf{Pass\textasciicircum{}4}
& \textbf{Pass\textasciicircum{}1} & \textbf{Pass\textasciicircum{}3} & \textbf{Pass\textasciicircum{}4}
& \textbf{Pass\textasciicircum{}1} & \textbf{Pass\textasciicircum{}3} & \textbf{Pass\textasciicircum{}4} \\
\midrule

Q3-32B-ReAct
& 33.0\% & 11.7\% & 8.2\%
& 29.6\% & 17.2\% & 15.2\%
& 36.1\% & 20.8\% & 16.7\%
& 28.0\% & 9.0\% & 6.0\% \\

Q3-32B-FC
& 35.0\% & 15.0\% & 13.1\%
& 17.6\% & 12.0\% & 10.8\%
& 32.0\% & 21.0\% & 18.0\%
& 42.0\% & 37.0\% & 35.0\% \\

Q3-32B-Thinking
& 34.1\% & 11.8\% & 8.2\%
& 43.0\% & 20.5\% & 16.0\%
& 36.7\% & 15.0\% & 12.2\%
& 29.0\% & 10.5\% & 7.0\% \\

\addlinespace[2pt]
\midrule
\addlinespace[2pt]

Base-4B
& 7.2\% & 6.3\% & 6.1\%
& \best{32.5\%} & \best{25.5\%} & \best{24.0\%}
& 25.0\% & 12.5\% & 11.1\%
& 22.5\% & 12.5\% & 10.0\% \\

RFT-4B
& 27.6\% & 14.3\% & 12.2\%
& 13.5\% & 8.5\% & 8.0\%
& \best{31.2\%} & \best{19.4\%} & 16.6\%
& 30.0\% & 22.5\% & 20.0\% \\

\oursrow
Policy-4B (\textbf{ours})
& 26.1\% & \best{17.4\%} & \best{16.5\%}
& 19.0\% & 12.5\% & 12.0\%
& 26.4\% & 16.7\% & \best{16.7\%}
& \best{33.8\%} & \best{30.0\%} & \best{30.0\%} \\

% \oursrow
% CATC-4B+CAST-4B (\textbf{ours})
% & \best{29.8\%} & 13.5\% & 10.4\%
% & 28.5\% & 14.5\% & 10.2\%
% & \best{33.3\%} & 9.7\% & 5.6\%
% & 28.8\% & 21.3\% & 20.0\% \\

\oursrow
Policy-4B+Critic-4B (\textbf{ours})
& \best{28.3\%} & 12.2\% & 9.6\%
& 27.5\% & 13.0\% & 10.0\%
& 29.2\% & 16.7\% & \best{16.7\%}
& 28.8\% & 21.3\% & 20.0\% \\

\addlinespace[2pt]
\midrule
\addlinespace[2pt]

Base-8B
& 14.1\% & 6.3\% & 5.2\%
& 13.5\% & 8.0\% & 8.0\%
& 31.9\% & 9.7\% & 5.6\%
& 32.5\% & 21.3\% & 20.0\% \\

RFT-8B
& 27.6\% & 13.9\% & 12.2\%
& 14.0\% & 7.5\% & 6.0\%
& 29.2\% & 12.5\% & 11.1\%
& \best{37.5\%} & \best{31.3\%} & \best{30.0\%} \\

\oursrow
Policy-8B (\textbf{ours})
& \best{30.0\%} & \best{16.7\%} & \best{14.8\%}
& 9.5\% & 6.5\% & 6.0\%
& 30.6\% & 12.5\% & 11.1\%
& \best{37.5\%} & 30.0\% & \best{30.0\%} \\

\oursrow
Policy-8B+Critic-8B (\textbf{ours})
& 29.8\% & 14.6\% & 13.0\%
& \best{20.5\%} & \best{13.0\%} & \best{12.0\%}
& \best{38.9\%} & \best{27.8\%} & \best{27.8\%}
& 32.5\% & 27.5\% & 25.0\% \\

\bottomrule
\end{tabular}
}
\caption{Comparison of CAST with RFT and base instruct models across four environments, including Retail, Airline, Telecom, and Telehealth. The best result within each model-size block is highlighted in yellow.}
\label{tab:tau_bench_results}
\end{table*}

CAST implements $V$ with multiple specialized LLM agents. A rule extractor agent retrieves relevant domain constraints and a tool extractor agent identifies candidate tools for the current step. Additional agents separately evaluate hallucination, domain violations, and tool usage. 
% To encourage strict grounding, the hallucination detection agent is isolated from $\Omega$ and auxiliary verifier outputs, so its judgment relies only on observable information at time step $t$. 
An orchestrator agent aggregates the outputs of agents and produces the final pair $(e_t,y_t)$.

\subsection{Critique Model Learning}

The critique model, CAST-Critic, is trained to generate both the verification rationale $e_t$ and the verification label $y_t$ from the action context
\[
C_\theta(h_t,o_t,a_t,\mathcal{K})
=
(\hat e_t,\hat y_t).
\]

We optimize $C_\theta$ using a combined objective
\[
\mathcal{L}_{\text{critique}}(\theta)
=
\lambda_{\text{exp}}
\mathcal{L}_{\text{exp}}(\theta)
+
\lambda_{\text{ver}}
\mathcal{L}_{\text{ver}}(\theta),
\]
where $\mathcal{L}_{\text{exp}}$ is the rationale generation loss and $\mathcal{L}_{\text{ver}}$ is the verification classification loss. This objective separates critique learning from policy optimization and provides direct supervision for producing both structured rationales and calibrated verification decisions.

\subsection{Critique-Aware Policy Optimization}

The trained CAST-Critic is then used to generate critique-enriched trajectories. Successful trajectories are selected to optimize the policy model, CAST-Policy.
At each step, CAST-Policy proposes an action
\[\bar a_t \sim \pi_\phi(\cdot \mid h_t,o_t,\mathcal{K}),\]
and CAST-Critic produces
\[(\hat e_t,\hat y_t)
=
C_\theta(h_t,o_t,\bar a_t,\mathcal{K}).\]
If the candidate action is accepted by CAST-Critic, we set \(a_t=\bar a_t\);
otherwise, the critique feedback is used to revise the action, and \(a_t\)
denotes the final accepted action. The final action and critique output are
then appended to the interaction context, forming a critique-enriched history
\[tilde h_{t+1} = h_t \oplus (a_t,\hat e_t,\hat y_t).\]
% \[
% \tilde h_t = h_t \oplus (a_t,\hat e_t,\hat y_t).
% \]
The interaction continues until task termination and is collected into a new trajectory buffer $\mathcal{B}_{C}$ containing observations, actions, final task rewards, and critique signals.

We select successful critique-enriched trajectories
\[
\mathcal{B}_{C}^{+}
=
\{\tau \in \mathcal{B}_{C} \mid Y(\tau)=1\},
\]
and optimize CAST-Policy on $\mathcal{B}_{C}^{+}$ using supervised fine-tuning to internalize verification signals during long-horizon tool use
\[
\mathcal{L}_{\text{policy}}(\phi)
=
-
\sum_{\tau \in \mathcal{B}_{C}^{+}}
\sum_{t=1}^{T}
\log \pi_\phi(a_t \mid \tilde h_t,o_t,\mathcal{K}).
\]

At inference time, CAST supports two modes. The optimized CAST-Policy can be used with CAST-Critic for explicit critique-guided interaction, or it can operate alone as an efficient standalone tool-calling agent.
\section{Experiments}
We evaluate CAST to answer three questions. First, does critique-aware training improve tool-calling reliability compared with base models and standard optimization baselines. Second, does CAST generalize from the in-domain training environment to unseen domains. Third, does the learned critique agent provide a more effective test-time verification signal than prompted frontier models and existing agentic frameworks. 

\begin{table}[!h]
\centering
\small
\setlength{\tabcolsep}{3.5pt}
\renewcommand{\arraystretch}{1.1}

\begin{tabular}{llccc}
\toprule
\textbf{Domain} 
& \textbf{Benchmark} 
& \shortstack{\textbf{No.}\\\textbf{Tools}}
& \shortstack{\textbf{No.}\\\textbf{Tasks}}
& \shortstack{\textbf{No. Turns}\\\textbf{(approx.)}} \\
\midrule

Airline
& $\tau$-bench
& 12
& 50
& [30--50] \\

Retail
& $\tau$-bench
& 14
& 114
& [30--50] \\

Telecom
& $\tau$-Trait
& 16
& 18
& [30--50] \\

Telehealth
& $\tau$-Trait
& 21
& 20
& [30--50] \\

\bottomrule
\end{tabular}

\caption{Overview of the evaluation benchmarks.}
\label{tab:overview_benchs}
\end{table}

\subsection{Experimental Setup}

\textbf{}{Models.}
We use Qwen3-4B and Qwen3-8B as the backbone models for both critique agents and critique-aware tool-calling agents. For each model size, we fine-tune one model to produce structured verification rationales and another model to act as a critique-aware policy. We compare these models with their corresponding base instruct models and with larger tool-calling models, including Qwen3-32B, Qwen3-235B-A22B-Instruct-2507~\cite{yang2025qwen3}, GPT-4.1~\cite{achiam2023gpt}, and GPT-OSS-120B~\cite{agarwal2025gpt}. Fine-tuning details and hyper-parameters are reported in Appendix~\ref{app:fine_tunning_detals}.

\textbf{Dataset.}
We construct the CAST training set from the official $\tau$-Bench Retail split, which contains 500 tasks. For each task, we sample 5 task trajectories with Qwen2.5-72B-Instruct as user model and Qwen3-32B as assistant model and apply CAST’s agentic verification procedure to annotate the resulting trajectories. This yields 2,500 critique-enriched experiences with structured step-level feedback for critique learning and policy optimization. Data generation details are in Appendix~\ref{app:dataset_details}.

% \begin{figure}[h]
%     \centering
%     \setlength{\fboxsep}{0pt} % Removes inner padding
%     % Creates an empty box 6cm wide and 4cm high
%     \fbox{\parbox[c][4cm][c]{6cm}{\centering (Placeholder Image)}} 
%     \caption{Talking about dataset}
%     \label{fig:empty_frame}
% \end{figure}

\textbf{Benchmarks.}
% We evaluate on $\tau$-Bench and $\tau$-Trait, covering four dynamic tool-calling domains including Retail, Airline, Telecom, and Telehealth. To test generalization, all fine-tuning is performed only on Retail. We therefore treat Retail as the in-domain environment and use Airline, Telecom, and Telehealth as out-of-domain environments. This setup evaluates whether CAST transfers critique-aware decision making beyond the domain used for training.
We evaluate all methods on $\tau$-Bench~\cite{yao2024tau} and $\tau$-Trait~\cite{he2025impatient} across four dynamic tool-calling domains, including Retail, Airline, Telecom, and Telehealth as summarized in Table~\ref{tab:overview_benchs}. All models are fine-tuned only on the Retail training split, which we use as the in-domain environment. Airline, Telecom, and Telehealth are reserved for out-of-domain evaluation to measure transfer beyond the training domain. We use pass\textasciicircum{}1 to measure task success in a single execution and pass\textasciicircum{}4 to measure reliability across repeated executions. Details of the pass\textasciicircum{}k estimator and benchmark protocol are provided in Appendix~\ref{app:details_benchmakrs}.

% \paragraph{Metrics.}
% We report pass\textasciicircum{}k for k from 1 to 4. For each task, pass\textasciicircum{}k estimates the probability that at least one of k independent executions succeeds. Given n sampled trajectories for a task and c successful trajectories, we compute pass\textasciicircum{}k using the standard unbiased estimator

% \begin{equation}
%     \text{Pass\textasciicircum{}k} = \mathbb{E}_{\text{tasks}} \left[ 1 - \frac{\binom{n-c}{k}}{\binom{n}{k}} \right]
% \end{equation}

% where C denotes the binomial coefficient. We use pass\textasciicircum{}1 to measure single-run task success and pass\textasciicircum{}4 to measure repeated-run reliability. Additional benchmark and metric details are provided in Appendix X.

\textbf{Baselines.}
We compare CAST with three groups of baselines. The first group includes standard prompting and tool-calling methods, including ReAct and function calling. The second group includes test-time agentic frameworks such as EvoTool, PALADIN, IRMA~\cite{mishra2025can} and FAMA~\cite{saeidi2026fama}, which use helper agents to support the main tool-calling agent during decision making. The third group includes optimization-based baselines, primarily rejection fine-tuning. In RFT, trajectories are collected, successful executions are selected according to task reward, and the model is optimized with supervised fine-tuning on the filtered data. This baseline isolates the effect of CAST’s critique supervision from simply training on successful trajectories.

\subsection{Main Results}

\begin{tcolorbox}[colback=green!5, colframe=green!50!black, breakable, left=1mm, right=1mm, top=1mm, bottom=1mm, boxsep=0mm]
% \small
\noindent \textbf{Key takeaway.} CAST-Policy consistently improves in-domain reliability across model sizes, with the strongest gains on pass\textasciicircum{}4.
\end{tcolorbox}

As shown in Table~\ref{tab:tau_bench_results}, on the Retail in-domain environment, CAST-Policy-4B improves over the base instruct model by 18.9\% on pass\textasciicircum{}1 and 10.4\% on pass\textasciicircum{}4. Although CAST-Policy-4B achieves comparable pass\textasciicircum{}1 performance to RFT-4B, it improves pass\textasciicircum{}4 by 4.3\%, showing that CAST-Policy improves repeated-run reliability beyond simply training on successful trajectories. CAST-Policy-8B shows the same trend, outperforming RFT-8B and Base-8B by 2.4\% and 15.9\% on pass\textasciicircum{}1, and by 2.6\% and 9.6\% on pass\textasciicircum{}4, respectively.

\begin{table}[t]
\centering
\small
\setlength{\tabcolsep}{6pt}
\begin{tabular}{l|cc}
\toprule
\textbf{Method} 
& \textbf{Avg Pass\textasciicircum{}1} 
& \textbf{Avg Pass\textasciicircum{}4} \\
\midrule

\multicolumn{3}{l}{\textit{\textbf{Qwen3-32B (Actor)}}} \\
\addlinespace[3pt]

\hspace{0.1em} + ReAct - no CA  
& 31.7\% & 11.5\% \\

\hspace{1em}+ CA (GPT-4.1)      
& 22.8\% & 6.0\% \\

\hspace{1em}+ CA (Q3-235B)    
& 24.1\% & 3.4\% \\

\hspace{1em}+ CA (Q2.5-72B)   
& 17.9\% & 6.3\% \\

\addlinespace[2pt]
\midrule
\addlinespace[2pt]

\oursrow
\hspace{1em}+ CA (Critic-4B)    
& \best{31.8\%} & 15.9\% \\

\oursrow
\hspace{1em}+ CA (Critic-8B)    
& 30.4\% & \best{19.5\%} \\

\bottomrule
\end{tabular}
\caption{Comparison of the CAST-Critic models with frontier models when used as critique agent (CA).}
\label{tab:critique_agent_results}
\end{table}

\begin{tcolorbox}[colback=green!5, colframe=green!50!black, breakable, left=1mm, right=1mm, top=1mm, bottom=1mm, boxsep=0mm]
% \small
\noindent \textbf{Key takeaway.} Smaller CAST-Policy models surpass Qwen3-32B on pass\textasciicircum{}4, showing stronger repeated-run reliability.
\end{tcolorbox}

Table~\ref{tab:tau_bench_results} also shows that CAST-Policy improves reliability-oriented performance relative to larger models. Although CAST-Policy-4B and CAST-Policy-8B have lower pass\textasciicircum{}1 performance than Qwen3-32B, they outperform the best Qwen3-32B setup by 3.4\% and 1.7\% on pass\textasciicircum{}4, respectively. This result shows that CAST-Policy can produce smaller agents that are more reliable across repeated executions than the larger model used during experience collection. This finding is further supported by Figure~\ref{fig:main_com}.
Appendix \ref{app:trajectory_case_study} provides qualitative examples that show how critique-aware agents can prevent premature irreversible actions from being executed when compared to a Qwen3-8B baseline model. 

\begin{tcolorbox}[colback=green!5, colframe=green!50!black, breakable, left=1mm, right=1mm, top=1mm, bottom=1mm, boxsep=0mm]
% \small
\noindent \textbf{Key takeaway.} Agentic CAST generalizes better to out-of-domain environments.
\end{tcolorbox}

In out-of-domain environments, agentic CAST shows stronger generalization than the standalone critique-aware policy. As shown in Table~\ref{tab:tau_bench_results}, on Airline and Telecom, CAST-Policy-8B paired with CAST-Critic-8B outperforms both the base instruct model and RFT-8B on pass\textasciicircum{}1 and pass\textasciicircum{}4. We also observe consistent gains in the remaining out-of-domain settings compared with competing methods. These results indicate that CAST’s critique-guided interaction transfers beyond the training domain and improves robustness under domain shift.

\begin{figure}[t]
    \centering
    \includegraphics[width=1\linewidth]{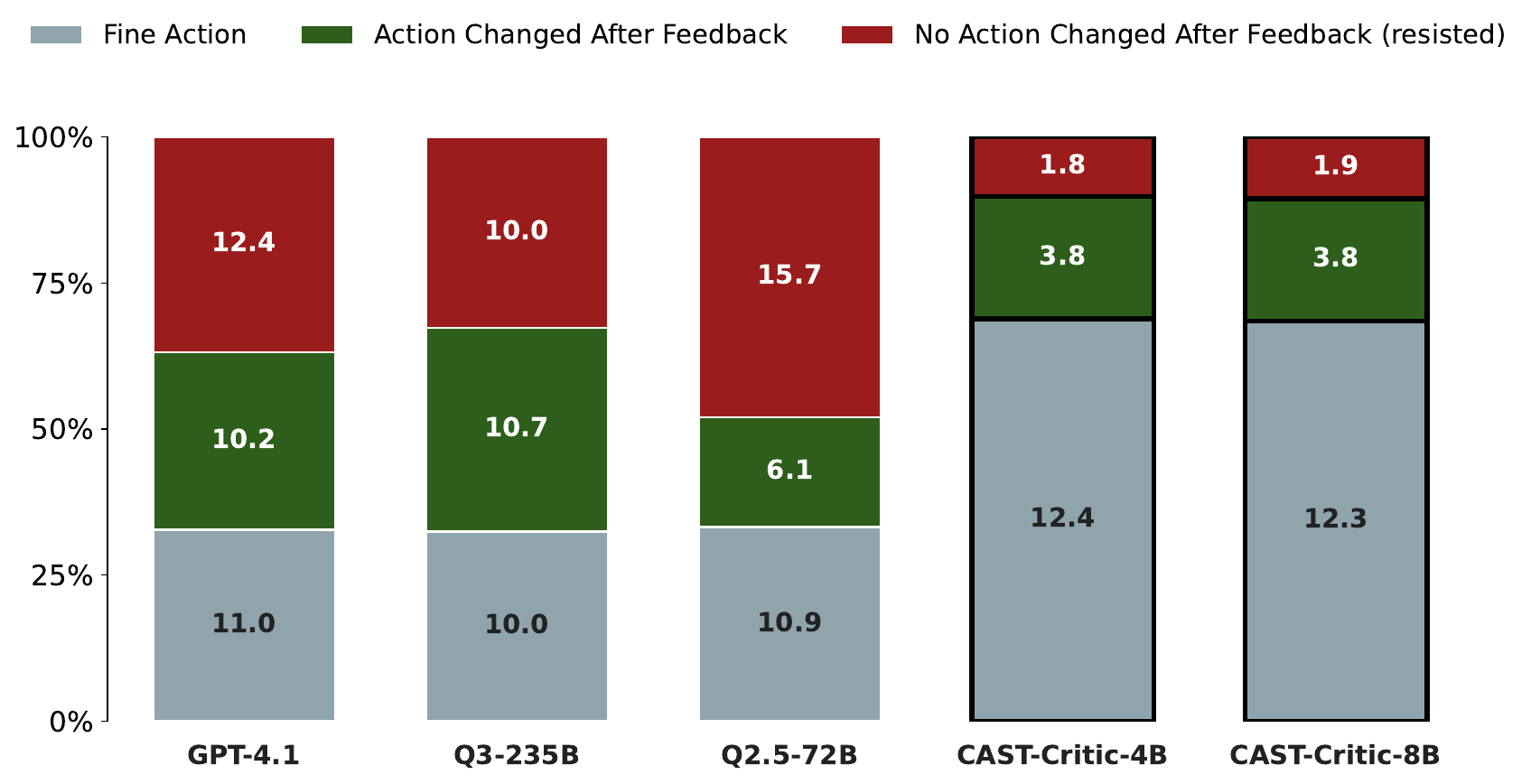}
    \caption{Action distribution of the tool-calling agent under different critique agents.}
    \label{fig:critique_only_comp}
\end{figure}

\begin{tcolorbox}[colback=green!5, colframe=green!50!black, breakable, left=1mm, right=1mm, top=1mm, bottom=1mm, boxsep=0mm]
% \small
\noindent \textbf{Key takeaway.} CAST-Critic provides more calibrated verification than frontier models.
\end{tcolorbox}
To evaluate CAST-Critic, we fix the tool-calling agent as Qwen3-32B and compare different critique models under the same prompting setup. Table~\ref{tab:critique_agent_results} shows that directly using frontier models as critique agents introduces substantial noise and reduces the performance of the base tool-calling agent. Our analysis shows that prompted instruct and frontier models tend to be overly pessimistic, frequently labeling valid actions as problematic. This behavior lengthens conversations and pushes the agent into unnecessary revision loops. In contrast, our 4B and 8B critique agents produce more calibrated judgments and trigger revisions mainly when the tool-calling action is unsuitable. The action-trigger ratios in Figure~\ref{fig:critique_only_comp} further support this finding.

\begin{figure}[h]
    \centering
    \includegraphics[width=1\linewidth]{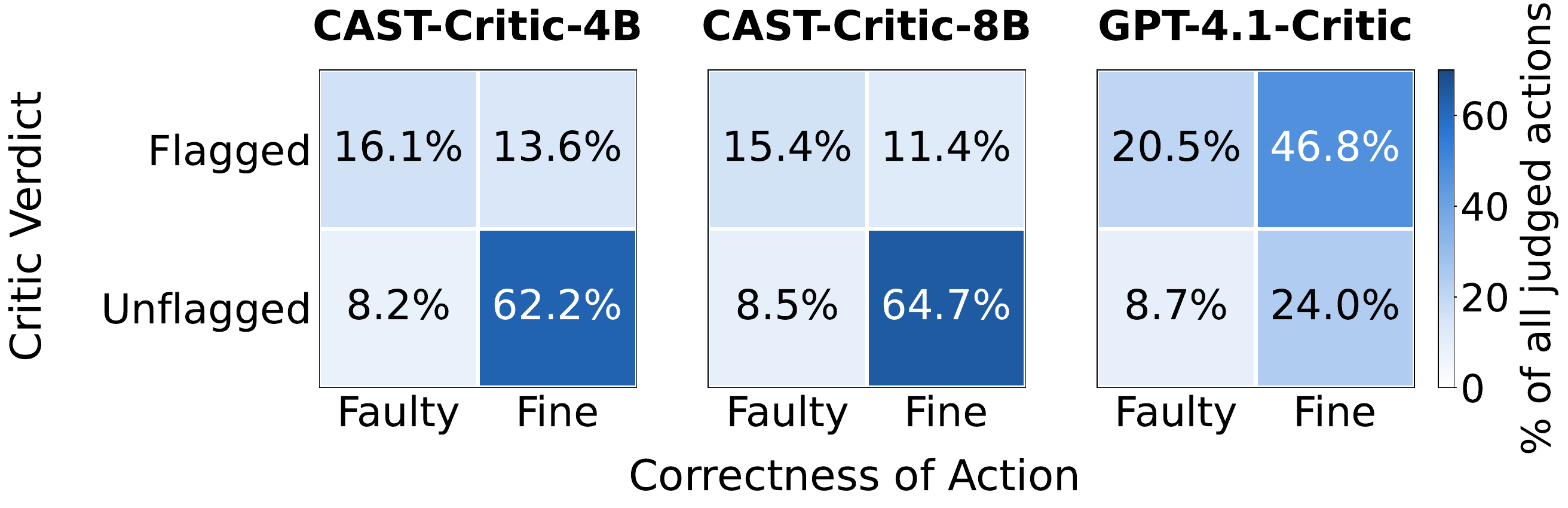}
    \caption{Confusion matrices of CAST-Critic-4B, CAST-Critic-8B, and GPT-4.1-Critic, comparing critic verdicts with action correctness.}
    \label{fig:confiusion_matrix}
\end{figure}

Consistent with this trend, the confusion matrices in Figure~\ref{fig:confiusion_matrix} provide a more detailed view of the critics' behavior. GPT-4.1-Critic flags 46.8\% of all judged actions even when they are correct, indicating a high false-positive rate and explaining the unnecessary revisions observed in Figure~\ref{fig:critique_only_comp}. In contrast, CAST-Critic-4B and CAST-Critic-8B reduce this rate to 13.6\% and 11.4\%, respectively, while maintaining comparable detection of faulty actions. At the same time, CAST-Critic-4B and CAST-Critic-8B correctly leave 62.2\% and 64.7\% of actions unflagged when they are valid, compared with only 24.0\% for GPT-4.1-Critic. These results further show that CAST-Critic better distinguishes genuinely faulty actions from valid ones, providing more calibrated verification and avoiding unnecessary intervention. 

\begin{figure}[t]
    \centering
    \includegraphics[width=1\linewidth]{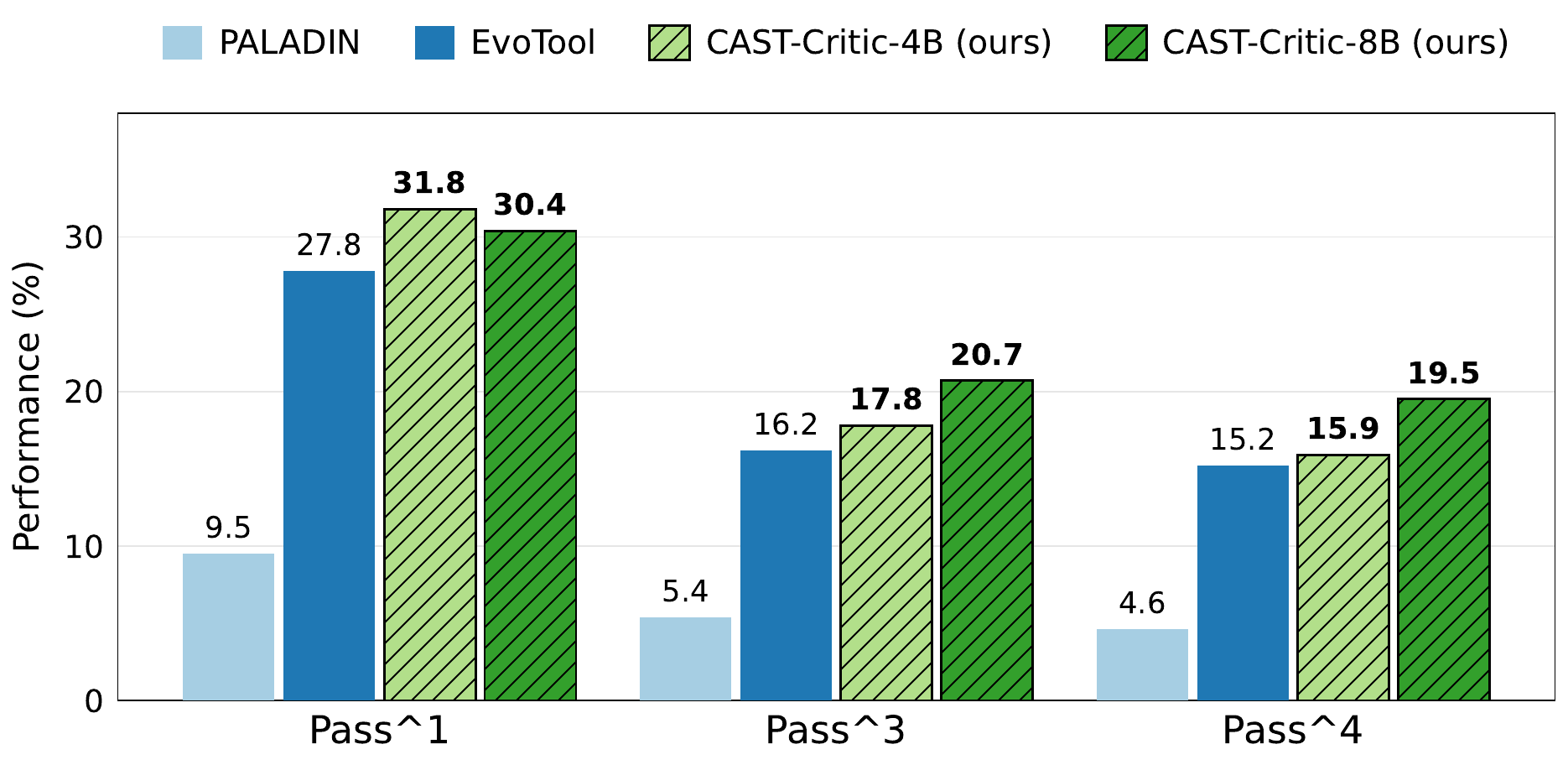}
    \caption{Comparison of CAST against PALADIN and EvoTool.}
    \label{fig:comp_evotool_paladin}
\end{figure}

\begin{figure*}[t]
    \centering
    \includegraphics[width=1\linewidth]{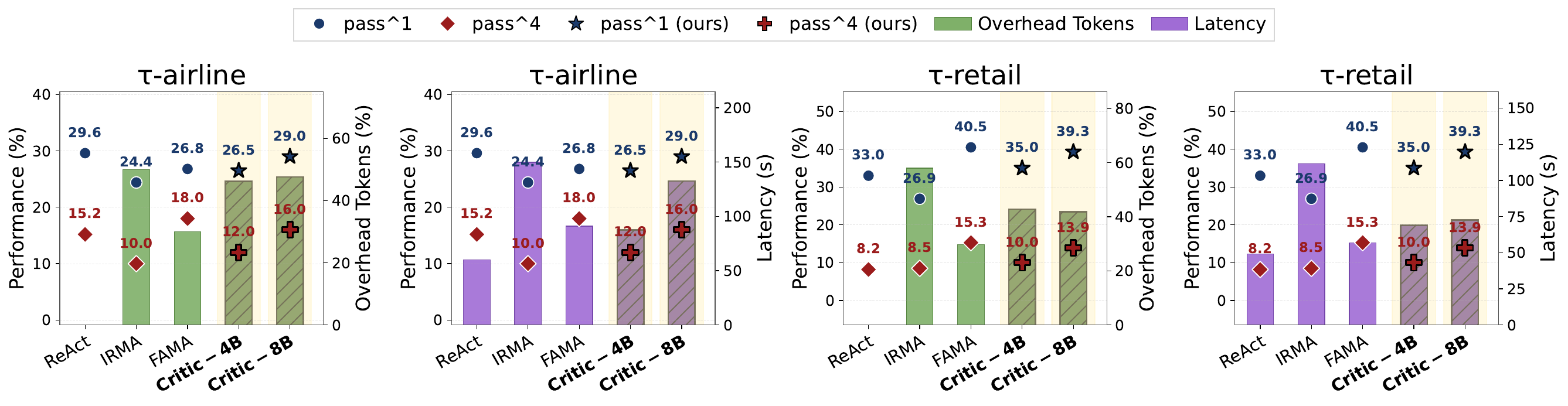}
    \caption{Comparison of CAST with agentic frameworks in terms of performance, token overhead, and latency.}
    \label{fig:agentic_comp}
\end{figure*}

\begin{figure}[h]
    \centering
    \includegraphics[width=1\linewidth]{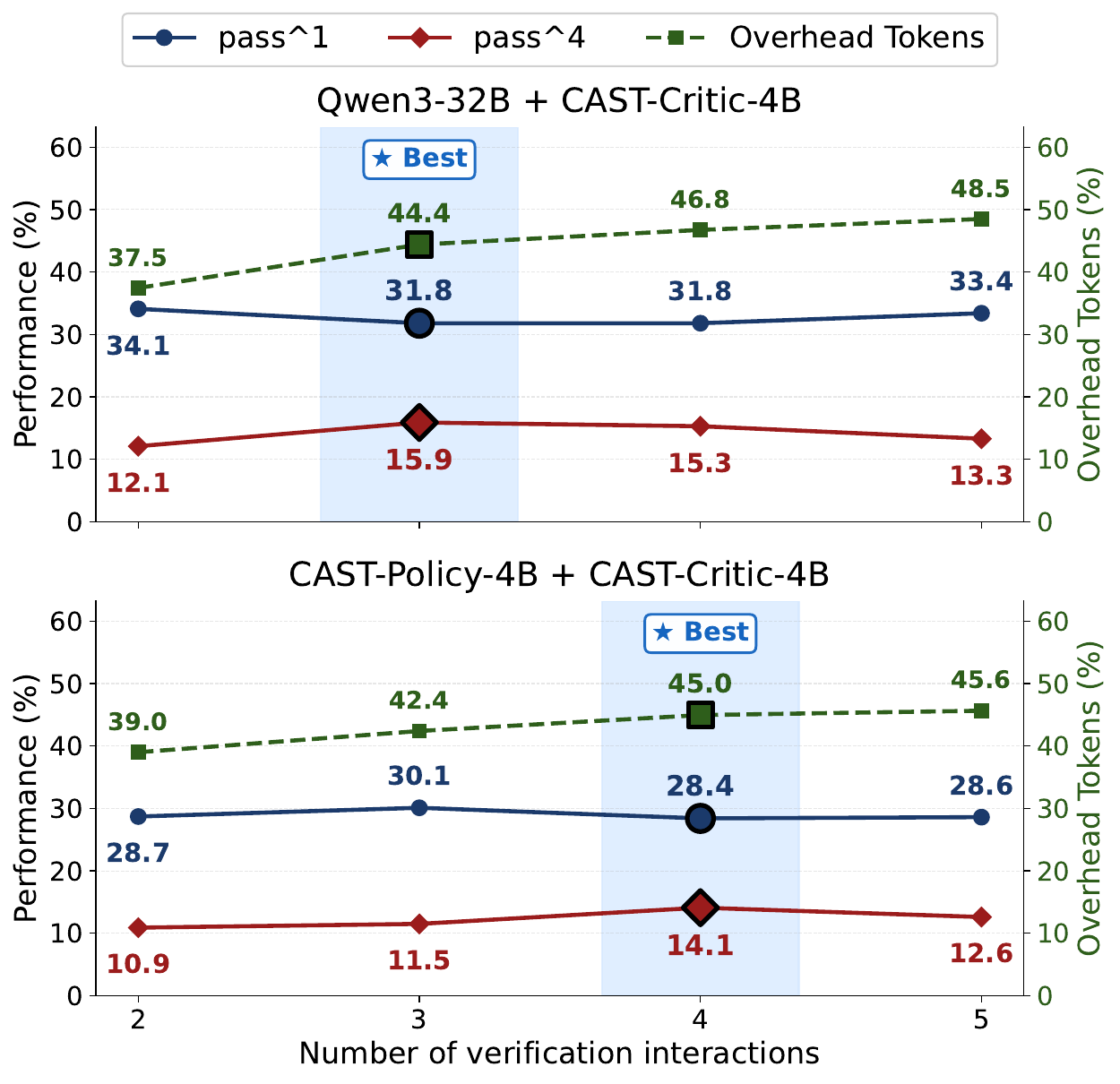}
    \caption{Comparison of performance and token overhead across different numbers of verification steps.}
    \label{fig:ablation_verification_steps}
\end{figure}

\begin{tcolorbox}[colback=green!5, colframe=green!50!black, breakable, left=1mm, right=1mm, top=1mm, bottom=1mm, boxsep=0mm]
% \small
\noindent \textbf{Key takeaway.} CAST achieves strong reliability against test-time agentic frameworks with smaller critique agents and moderate overhead.
\end{tcolorbox}

We compare CAST with test-time reliability-oriented frameworks such as IRMA and FAMA, using Qwen3-32B as the tool-calling agent across all methods. As shown in Figure~\ref{fig:agentic_comp}, CAST outperforms IRMA on Airline by 3.0\% on pass\textasciicircum{}1 and on Retail by 4.5\% on pass\textasciicircum{}4, while using 4B or 8B critique agents instead of the Qwen2.5-72B helper agents used by IRMA and FAMA. CAST introduces more overhead than single-pass inference because it performs critique-guided refinement, but its token usage and latency remain within the range of existing agentic frameworks. These results show that CAST can achieve competitive reliability with substantially smaller helper models and practical inference cost.

We further compare CAST with PALADIN and EvoTool, two methods designed to improve the reliability of tool-using agents. PALADIN improves failure recovery by training on recovery-annotated trajectories with injected tool failures and retrieving similar failure cases at inference time, whereas EvoTool iteratively evolves a modular tool-use policy through blame-aware mutation and diversity-aware selection. As shown in Figure~\ref{fig:comp_evotool_paladin}, CAST substantially outperforms PALADIN across all evaluation settings, with CAST-Critic-8B improving pass\textasciicircum{}1, pass\textasciicircum{}3, and pass\textasciicircum{}4 by 20.9, 15.3, and 14.9 percentage points, respectively. Compared with EvoTool, CAST-Critic-8B achieves improvements of 2.6, 4.5, and 4.3 percentage points on pass\textasciicircum{}1, pass\textasciicircum{}3, and pass\textasciicircum{}4, respectively. These results demonstrate that a simple actor--critic formulation with CAST can outperform both failure-recovery and evolutionary tool-use methods.

\begin{figure*}[t]
    \centering
    \includegraphics[width=1\linewidth]{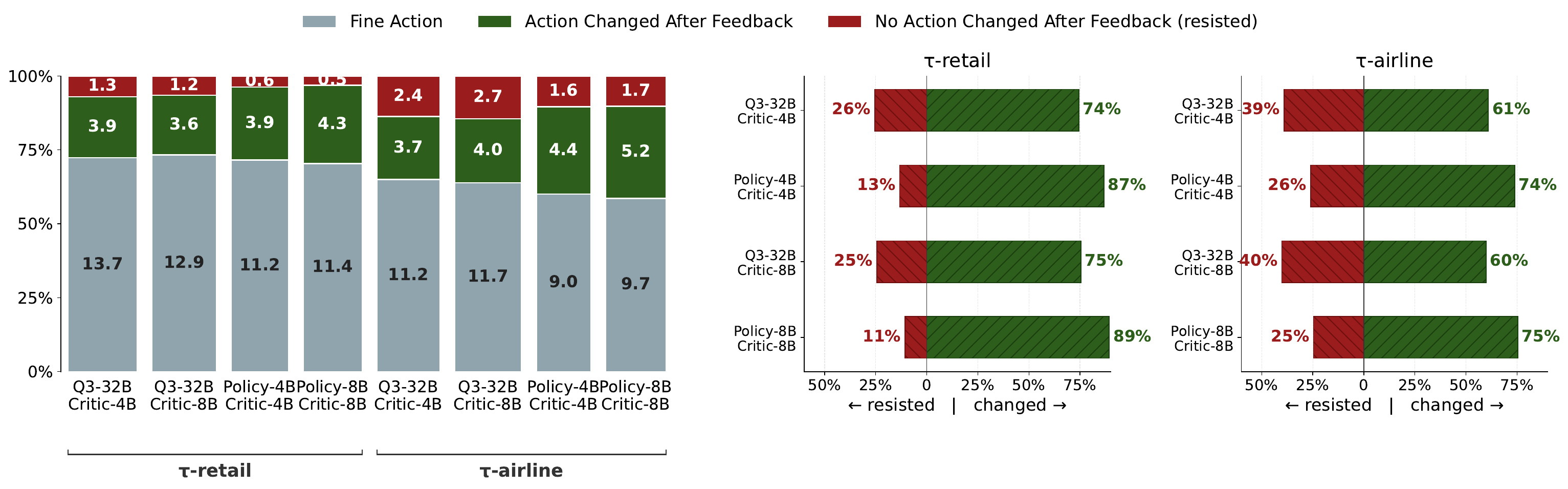}
    \caption{\textbf{Left:} Distribution of tool-calling agent actions across different agentic scenarios. \textbf{Right:} Comparison of how often instruct and CATC agents resist the verification signal after receiving critique feedback.}
    \label{fig:resistant_analysis}
\end{figure*}

\begin{figure}[h]
    \centering
    \includegraphics[width=1\linewidth]{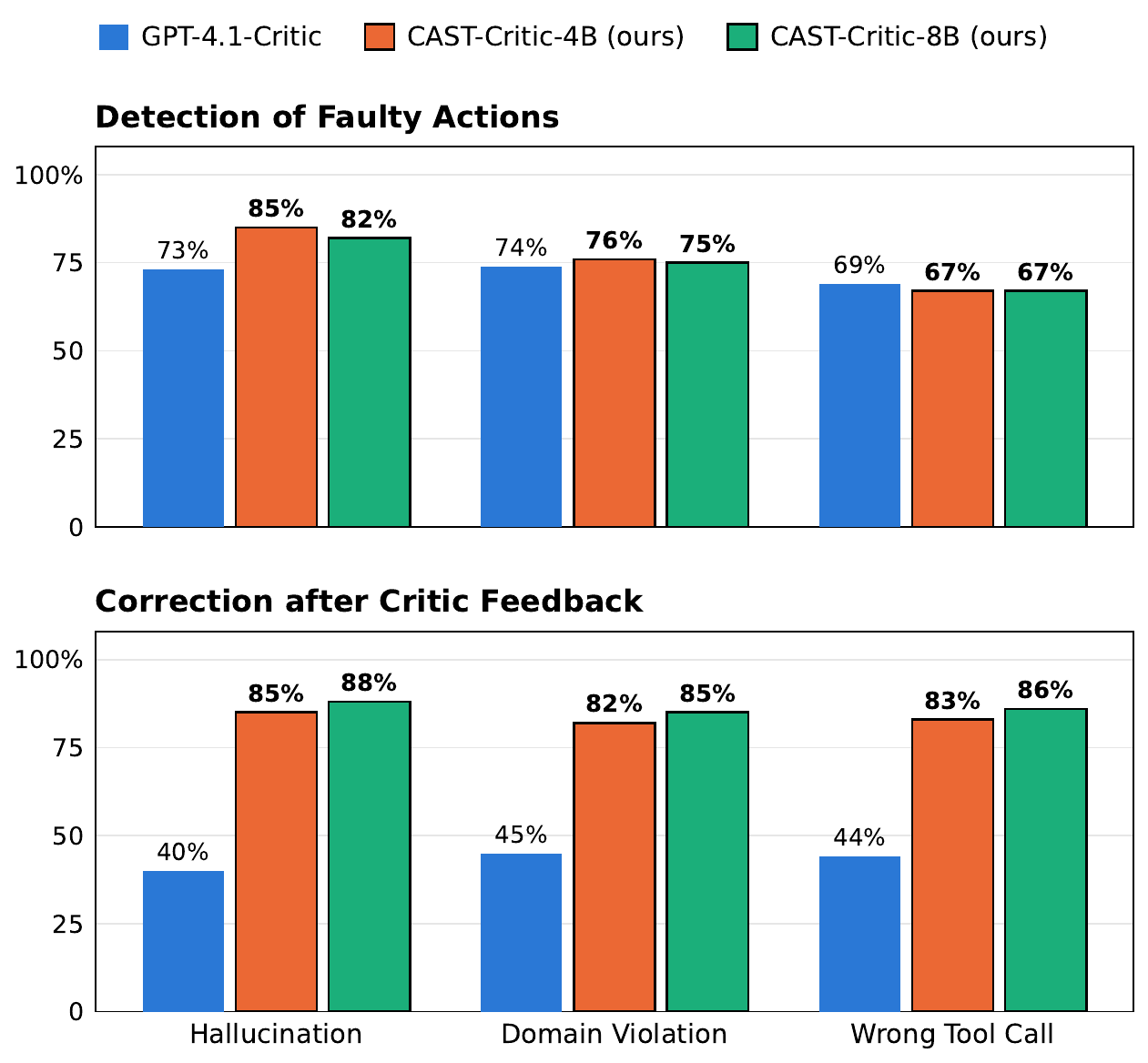}
    \caption{\textbf{Left:} critic accuracy in detecting faulty actions across hallucination, domain violation, and wrong tool-call. \textbf{Right:} correction rate after critic feedback for the same error types. The actor model is Qwen3-32B.}
    \label{fig:critic_error_analysis}
\end{figure}

\subsection{Ablation and Analysis}

\begin{tcolorbox}[colback=blue!5, colframe=blue!50!black, breakable, left=1mm, right=1mm, top=1mm, bottom=1mm, boxsep=0mm]
% \small
\noindent \textbf{Key takeaway.} The best verification budget depends on the tool-calling agent.
\end{tcolorbox}
To study the effect of verification budget, we vary the number of refinement rounds from 2 to 5 after the critique agent flags an unsuitable action. The ablation uses two tool-calling agents, Qwen3-32B and CAST-Policy-4B, with CAST-Critic-4B fixed as the critique model. As shown in Figure~\ref{fig:ablation_verification_steps}, additional verification increases token overhead for both agents, but reliability does not improve monotonically. Qwen3-32B achieves its best pass\textasciicircum{}4 with 3 rounds, whereas CAST-Policy-4B reaches its best pass\textasciicircum{}4 with 4 rounds. This indicates that the optimal verification budget depends on how the tool-calling agent responds to critique feedback. In almost all settings, the CAST-Policy-4B uses fewer tokens than the Qwen3-32B, suggesting that the critique-aware policy incorporates verification feedback with lower interaction cost.

\begin{tcolorbox}[colback=blue!5, colframe=blue!50!black, breakable, left=1mm, right=1mm, top=1mm, bottom=1mm, boxsep=0mm]
% \small
\noindent \textbf{Key takeaway.} CAST-Critic identifies problematic actions without over-triggering revisions. 
\end{tcolorbox}
We analyze test-time verification behavior under two settings. In the first setting, Qwen3-32B is used as the tool-calling agent and paired with our 4B or 8B critique agent. In the second setting, CAST-Policy-4B and CAST-Policy-8B are paired with CAST-Critic of the same scale. The left plot in Figure~\ref{fig:resistant_analysis} shows the distribution of verification outcomes across these settings. Our critique agents classify most tool-calling actions as suitable and flag only 33.2\% of agent actions as unsuitable per task. This supports the finding that the learned critique agents are not overly pessimistic and do not frequently push the tool-calling agent into unnecessary revision loops.

\begin{tcolorbox}[colback=blue!5, colframe=blue!50!black, breakable, left=1mm, right=1mm, top=1mm, bottom=1mm, boxsep=0mm]
% \small
\noindent \textbf{Key takeaway.} CAST-Policy models are less resistant to verification feedback than non-fine-tuned tool-calling agents.
\end{tcolorbox}
We further examine how the tool-calling agent responds after receiving critique feedback. We distinguish between cases where the agent accepts the critique and revises its action and cases where it resists the critique and keeps the original action. As shown in the right plot of Figure~\ref{fig:resistant_analysis}, CAST-Policy agents exhibit roughly 2 times less resistance to verification feedback than non-fine-tuned agents. This explains why smaller CAST-Policy models can achieve performance comparable to larger non-fine-tuned tool-calling agents when paired with our critique agents. Together with the pass\textasciicircum{}4 results in Tables~\ref{tab:tau_bench_results} and ~\ref{tab:critique_agent_results}, this indicates that CAST improves not only the quality of the critique signal but also the policy’s ability to act on that signal.

% \begin{tcolorbox}[colback=blue!5, colframe=blue!50!black, breakable, left=1mm, right=1mm, top=1mm, bottom=1mm, boxsep=0mm]
% % \small
% \noindent \textbf{Key takeaway.} CAST Critics produce richer verification feedback.
% \end{tcolorbox}

% In Table~\ref{tab:critique_agent_results} we have shwed the CAST critics outperform frontier open source and close source. We dig into the outputs and did a comperhensive analysis on the failure detection and effectivness of the feedback of CAST critics and GPT-4.1 as critic agent. The results in Figure~\ref{} show that CAST critics have impresive performance to detect halluciantion and domain violation. The results also show that they comparable performance on detecting Wrong Tool Calling errors. However we in Figure~\ref{} Right we can see the how many of the feedback cause to solve the issue with a proper explanation and in all error categories the feedback produced by CAST critics can solve the issue by 88\% that shows the quality and vision of CAST critics. 

\begin{tcolorbox}[colback=blue!5, colframe=blue!50!black, breakable, left=1mm, right=1mm, top=1mm, bottom=1mm, boxsep=0mm]
% \small
\noindent \textbf{Key takeaway.} CAST-Critics provide more effective verification and corrective feedback.
\end{tcolorbox}

Table~\ref{tab:critique_agent_results} shows that CAST Critics outperform strong open- and closed-source critic models. To better understand these gains, we conduct a fine-grained analysis of their ability to detect different failure types and the effectiveness of their feedback in correcting faulty actions, using GPT-4.1 as the baseline critic. As shown in Figure~\ref{fig:critic_error_analysis} (left), CAST Critics substantially outperform GPT-4.1 in detecting hallucinations, achieve comparable or slightly better performance on domain violations, and remain competitive on wrong tool calls. More importantly, Figure~\ref{fig:critic_error_analysis} (right) shows that feedback from CAST Critics leads to successful correction in 82--88\% of cases across all three error categories, substantially outperforming GPT-4.1. These results highlight that CAST Critics not only identify erroneous actions effectively but also provide actionable feedback that helps the actor correct them.

\section{Conclusion}
% We introduced CAST, a critique-aware training framework for improving the reliability of tool-calling agents in dynamic long-horizon environments. CAST addresses the challenge that a single unreliable action can derail an otherwise promising interaction by learning structured action-level verification rationales and using them for both critique learning and policy optimization. Instead of relying only on prompted frontier models as test-time critics, CAST constructs verification supervision through agentic trajectory analysis and trains dedicated critique agents that provide more calibrated feedback. Our experiments show that CAST improves reliability across three usage modes, including standalone critique-aware policies, critique-guided inference with large tool-calling models, and the full agentic setting where the critique agent and policy agent interact. Across in-domain and out-of-domain environments, CAST improves task success and repeated-run reliability while using substantially smaller critique agents than existing test-time agentic frameworks. Our analysis further shows that trained critique agents avoid the overly conservative behavior of prompted frontier models and help tool-calling agents revise problematic actions without falling into unnecessary revision loops. Overall, these results suggest that structured critique supervision is a scalable path toward LLM agents that can reason about intermediate actions, recover from unreliable decisions, and complete complex tasks more consistently in realistic dynamic environments.

We presented CAST, a critique-aware framework for making tool-calling agents more reliable in dynamic long-horizon interactions. CAST starts from the observation that many failures arise from intermediate actions that appear plausible but later derail the task. To address this, CAST generates structured verification rationales for agent actions, trains a critique agent to provide calibrated feedback, and uses this feedback to optimize a critique-aware tool-calling policy. Experiments across in-domain and out-of-domain settings show that this training process improves both task success and repeated-run reliability. These results suggest that learning to critique intermediate actions is a practical path toward agents that solve complex tasks more consistently.

\section*{Limitations}
% Although CAST improves the reliability of tool-calling agents, two limitations remain. First, CAST relies on supervised fine-tuning for both critique learning and policy optimization. While this provides stable training, it does not directly optimize the policy through on-policy interaction with critique feedback. Future work could use the learned critique agent as a feedback model for reinforcement learning, enabling the policy to learn more adaptive correction strategies. Second, CAST focuses on verifying the current policy action rather than modeling its long-term consequences. Extending critique supervision with forward-looking rationales could help agents better understand how local decisions affect later task outcomes in long-horizon interactions.
Although CAST improves the reliability of tool-calling agents, two limitations remain. First, CAST trains the critique agent and policy agent with supervised fine-tuning, which provides stable optimization but does not directly train the policy through on-policy interaction with critique feedback. Future work could use the learned critique agent as a feedback model for reinforcement learning, allowing the policy to learn more adaptive correction strategies from its own exploration. Second, CAST focuses on verifying the current policy action rather than explicitly modeling its downstream consequences. Since CAST generates critique data during training, future extensions could incorporate forward-looking rationales that describe how a local action affects later states or task outcomes, providing stronger supervision for long-horizon decision making.

\section*{Ethics Statement}
We have utilized AI assistants, specifically Grammarly and ChatGPT, to correct grammatical errors and rephrase sentences.

\section*{Acknowledgement}

We thank the anonymous reviewers for their constructive suggestions. We extend our gratitude to Research Computing (RC) and Enterprise Technology at ASU for providing computing resources and access to OpenAI models for experiments. This work was in part supported by a gift award from Cisco Research. Baral was also partially supported by a DARPA contract.

\bibliography{custom}
\clearpage
\appendix

\section{Details about Fine Tuning}
\label{app:fine_tunning_detals}
We fine-tune Qwen-series models using the multi-turn agentic data we generated on one nodes of a supercomputing cluster equipped with NVIDIA H100 GPUs. The maximum sequence length is set to 32,768 for Qwen3 models. Detailed hyperparameters are provided in Table~\ref{tab:hyperparameters}.
\begin{table}[h]
\centering
\begin{tabular}{l c}
\toprule
\textbf{Hyper-parameter} & \textbf{Value} \\
\midrule
Learning Rate & $5 \times 10^{-6}$ \\
Number of Epochs & 3 \\
Number of Devices & 4 \\
Total Batch Size & 16 \\
Optimizer & AdamW \\
Learning Rate Scheduler & cosine \\
\bottomrule
\end{tabular}
\caption{Fine-tuning hyperparameters used for Qwen3 models.}
\label{tab:hyperparameters}
\end{table}

\section{Details about Dataset}
\label{app:dataset_details}
We fine-tune two models in CAST, a critique model and a policy model. To construct the initial trajectory data, we use Qwen3-32B in ReAct mode as the teacher policy. We use the official $\tau$-Bench training split, which contains 500 tasks, and run the teacher model five times per task with temperature 0.7 to collect diverse solution trajectories. In these rollouts, Qwen2.5-72B is used as the user simulator. After collecting the trajectories, we annotate each policy action using our agentic verification framework. We use Qwen3.5-27B and Qwen3.5-122B-A10B as verification agents and find that Qwen3.5-122B-A10B produces higher-quality annotations. The annotated trajectories are then reconstructed into multi-turn training examples for the critique model, with the tool list and domain policy provided in the system prompt to improve generalization.

After fine-tuning the critique model, we perform a second round of training data collection. We again run the 500 $\tau$-Bench training tasks for five trials, but now the teacher policy actions are verified by the learned critique model during interaction. We retain only trajectories with reward 1 and use these successful critique-enriched trajectories to train the critique-aware tool-calling policy.

\section{Details about Benchmarks}
\label{app:details_benchmakrs}
\paragraph{$\tau$-bench}
$\tau$-bench evaluates tool-using conversational agents in realistic customer-service environments. The benchmark includes two domains, Retail and Airline, where an agent interacts with a simulated user, follows domain-specific policy rules, and uses API tools to query or update a backend state. The Retail domain contains 500 users, 50 products, and 1,000 orders, with 15 tools in total, including 7 write tools and 8 non-write tools. It includes 115 tasks covering cancellations, returns, exchanges, address changes, and information requests. The Airline domain contains 500 users, 300 flights, and 2,000 reservations, with 13 tools in total, including 6 write tools and 7 non-write tools. It includes 50 tasks covering reservation booking, modification, cancellation, and refund-related workflows.

\paragraph{$\tau$-trait}
$\tau$-trait extends $\tau$-bench to evaluate robustness under more diverse user behaviors and domain shifts. It preserves the multi-turn tool-use setting while introducing persona-aware user simulation and expanding the benchmark beyond Retail and Airline to include Telecom and Telehealth. The benchmark is designed to test robustness, personalization, and fairness under realistic behavior variation. It contains 218 tasks across four domains, including Retail with 120 tasks, Airline with 60 tasks, Telecom with 18 tasks, and Telehealth with 20 tasks.

For both $\tau$-bench and $\tau$-trait, we evaluate task success using the pass\textasciicircum{}$k$ metric~\citep{yao2024tau}. This metric measures the probability that $k$ independently sampled executions of a task are all successful, averaged across tasks. Given $n$ independent trials for a task and $c$ successful executions with reward $r=1$, pass\textasciicircum{}$k$ is computed as
\begin{equation*}
    \text{pass\textasciicircum{}$k$}
    =
    \mathbb{E}_{\text{task}}
    \left[
    \frac{\binom{c}{k}}{\binom{n}{k}}
    \right].
\end{equation*}

\section{Details Results}
We report the detailed results of all experiments conducted in this study in Tables 4–10.

% \subsection{More Analysis on Critics}

\begin{table*}[t]
    \centering
    \scriptsize
    \setlength{\tabcolsep}{3pt}
    \resizebox{\textwidth}{!}{
        \begin{tabular}{l | c c c c | c c c c}
            \toprule
            \multirow{2}{*}{\textbf{Model}}                   &
            \multicolumn{4}{c|}{\textbf{$\tau$-Bench Retail}} &
            \multicolumn{4}{c}{\textbf{$\tau$-Bench Airline}}                                                                                                                                                                \\
            \cmidrule(lr){2-5}\cmidrule(lr){6-9}
                                                              &
            \textbf{Pass\textasciicircum{}1}                  & \textbf{Pass\textasciicircum{}2} & \textbf{Pass\textasciicircum{}3} & \textbf{Pass\textasciicircum{}4} &
            \textbf{Pass\textasciicircum{}1}                  & \textbf{Pass\textasciicircum{}2} & \textbf{Pass\textasciicircum{}3} & \textbf{Pass\textasciicircum{}4}                                                       \\
            \midrule

            GPT-OSS-120B    & 31.3\% & 16.2\% & 9.7\% & 5.9\% & 37.2\% & 28.4\% & 24.6\% & 22.0\% \\

            % \addlinespace[2pt]
            % \midrule
            % \addlinespace[2pt]

            Q3-235B         & 49.4\% & 36.8\% & 30.9\% & 27.8\% & 34.0\% & 25.3\% & 23.0\% & 22.0\% \\

            % \addlinespace[2pt]
            % \midrule
            % \addlinespace[2pt]

            Q3-32B-Thinking & 34.1\% & 18.4\% & 11.8\% & 8.2\%  & 43.0\% & 27.7\% & 20.5\% & 16.0\% \\
            Q3-32B-ReAct          & 33.0\% & 18.2\% & 11.7\% & 8.2\%  & 29.6\% & 21.0\% & 17.2\% & 15.2\% \\
            Q3-32B-FC             & 35.0\% & 26.6\% & 15.0\% & 13.1\% & 17.6\% & 13.6\% & 12.0\% & 10.8\% \\

            \bottomrule
        \end{tabular}
    }
    \caption{Results on $\tau$-Bench.}
    \label{tab:tau_bench_retail_airline}
\end{table*}

\begin{table*}[t]
    \centering
    \scriptsize
    \setlength{\tabcolsep}{3pt}
    \resizebox{\textwidth}{!}{
        \begin{tabular}{l | c c c c | c c c c}
            \toprule
            \multirow{2}{*}{\textbf{Model}}                    &
            \multicolumn{4}{c|}{\textbf{$\tau$-Trait Telecom}} &
            \multicolumn{4}{c}{\textbf{$\tau$-Trait Telehealth}}                                                                                                                                                              \\
            \cmidrule(lr){2-5}\cmidrule(lr){6-9}
                                                               &
            \textbf{Pass\textasciicircum{}1}                   & \textbf{Pass\textasciicircum{}2} & \textbf{Pass\textasciicircum{}3} & \textbf{Pass\textasciicircum{}4} &
            \textbf{Pass\textasciicircum{}1}                   & \textbf{Pass\textasciicircum{}2} & \textbf{Pass\textasciicircum{}3} & \textbf{Pass\textasciicircum{}4}                                                       \\
            \midrule

            GPT-OSS-120B    & 42.2\% & 30.6\% & 23.3\% & 18.9\% & 32.0\% & 26.0\% & 23.0\% & 21.0\% \\

            % \addlinespace[2pt]
            % \midrule
            % \addlinespace[2pt]

            Q3-235B         & 52.8\% & 42.6\% & 37.5\% & 33.3\% & 42.5\% & 35.0\% & 30.0\% & 25.0\% \\

            % \addlinespace[2pt]
            % \midrule
            % \addlinespace[2pt]

            Q3-32B-Thinking & 36.7\% & 20.6\% & 15.0\% & 12.2\% & 29.0\% & 17.0\% & 10.5\% & 7.0\%  \\
            Q3-32B-ReAct          & 36.1\% & 25.9\% & 20.8\% & 16.7\% & 28.0\% & 15.0\% & 9.0\%  & 6.0\%  \\
            Q3-32B-FC             & 32.0\% & 25.0\% & 21.0\% & 18.0\% & 42.0\% & 39.0\% & 37.0\% & 35.0\% \\

            \bottomrule
        \end{tabular}
    }
    \caption{Results on $\tau$-Trait.}
    \label{tab:tau_trait_telecom_telehealth}
\end{table*}

\begin{table*}[t]
    \centering
    \scriptsize
    \setlength{\tabcolsep}{3pt}
    \resizebox{\textwidth}{!}{
        \begin{tabular}{l | c c c c | c c c c}
            \toprule
            \multirow{2}{*}{\textbf{Model}}                   &
            \multicolumn{4}{c|}{\textbf{$\tau$-Bench Retail}} &
            \multicolumn{4}{c}{\textbf{$\tau$-Bench Airline}}                                                                                                                                                                \\
            \cmidrule(lr){2-5}\cmidrule(lr){6-9}
                                                              &
            \textbf{Pass\textasciicircum{}1}                  & \textbf{Pass\textasciicircum{}2} & \textbf{Pass\textasciicircum{}3} & \textbf{Pass\textasciicircum{}4} &
            \textbf{Pass\textasciicircum{}1}                  & \textbf{Pass\textasciicircum{}2} & \textbf{Pass\textasciicircum{}3} & \textbf{Pass\textasciicircum{}4}                                                       \\
            \midrule

            Q3-32B-FAMA & 40.0\% & 26.9\% & 19.9\% & 15.3\% & 26.8\% & 20.0\% & 18.4\% & 18.0\% \\
            Q3-32B-IRMA & 26.9\% & 15.2\% & 10.9\% & 8.5\%  & 24.4\% & 15.9\% & 12.4\% & 10.0\% \\

            \bottomrule
        \end{tabular}
    }
    \caption{Results on $\tau$-Bench.}
    \label{tab:tau_bench_cast_retail_airline}
\end{table*}

\begin{table*}[t]
    \centering
    \scriptsize
    \setlength{\tabcolsep}{3pt}
    \resizebox{\textwidth}{!}{
        \begin{tabular}{l | c c c c | c c c c}
            \toprule
            \multirow{2}{*}{\textbf{Model}}                    &
            \multicolumn{4}{c|}{\textbf{$\tau$-Trait Telecom}} &
            \multicolumn{4}{c}{\textbf{$\tau$-Trait Telehealth}}                                                                                                                                                              \\
            \cmidrule(lr){2-5}\cmidrule(lr){6-9}
                                                               &
            \textbf{Pass\textasciicircum{}1}                   & \textbf{Pass\textasciicircum{}2} & \textbf{Pass\textasciicircum{}3} & \textbf{Pass\textasciicircum{}4} &
            \textbf{Pass\textasciicircum{}1}                   & \textbf{Pass\textasciicircum{}2} & \textbf{Pass\textasciicircum{}3} & \textbf{Pass\textasciicircum{}4}                                                       \\
            \midrule

            Q3-32B-FAMA & 37.8\% & 31.7\% & 30.0\% & 28.9\% & 30.0\% & 24.5\% & 20.0\% & 18.0\% \\
            Q3-32B-IRMA & 40.0\% & 28.0\% & 24.4\% & 23.3\% & 45.0\% & 37.0\% & 35.0\% & 35.0\% \\

            \bottomrule
        \end{tabular}
    }
    \caption{Results on $\tau$-Trait.}
    \label{tab:tau_trait_cast_telecom_telehealth}
\end{table*}

\begin{table*}[t]
    \centering
    \scriptsize
    \setlength{\tabcolsep}{3pt}
    \resizebox{\textwidth}{!}{
        \begin{tabular}{l | c c c c | c c c c}
            \toprule
            \multirow{2}{*}{\textbf{Model}}                   &
            \multicolumn{4}{c|}{\textbf{$\tau$-Bench Retail}} &
            \multicolumn{4}{c}{\textbf{$\tau$-Bench Airline}}                                                                                                                                                                \\
            \cmidrule(lr){2-5}\cmidrule(lr){6-9}
                                                              &
            \textbf{Pass\textasciicircum{}1}                  & \textbf{Pass\textasciicircum{}2} & \textbf{Pass\textasciicircum{}3} & \textbf{Pass\textasciicircum{}4} &
            \textbf{Pass\textasciicircum{}1}                  & \textbf{Pass\textasciicircum{}2} & \textbf{Pass\textasciicircum{}3} & \textbf{Pass\textasciicircum{}4}                                                       \\
            \midrule

            Q3-32B+GPT-4.1  & 17.0\% & 7.5\%  & 4.3\% & 2.6\% & 29.0\% & 21.3\% & 18.0\% & 16.0\% \\
            Q3-32B+Q3-235B   & 20.9\% & 7.1\%  & 3.3\% & 1.7\% & 30.5\% & 21.3\% & 16.0\% & 12.0\% \\
            Q3-32B+Q2.5-72B  & 19.1\% & 10.6\% & 7.6\% & 6.1\% & 29.5\% & 22.0\% & 17.5\% & 14.0\% \\
            
            \addlinespace[2pt]
            \midrule
            \addlinespace[2pt]
            
            Q3-32B+CAST-Critic-4B & 35.0\% & 20.9\% & 13.9\% & 10.0\% & 26.5\% & 16.0\% & 13.0\% & 12.0\% \\
            Q3-32B+CAST-Critic-8B & 29.8\% & 18.4\% & 14.6\% & 13.0\% & 20.5\% & 15.0\% & 13.0\% & 12.0\%\\
            
            \bottomrule
        \end{tabular}
    }
    \caption{Results on $\tau$-Bench.}
    \label{tab:tau_bench_loop3_retail_airline}
\end{table*}

\begin{table*}[t]
    \centering
    \scriptsize
    \setlength{\tabcolsep}{3pt}
    \resizebox{\textwidth}{!}{
        \begin{tabular}{l | c c c c | c c c c}
            \toprule
            \multirow{2}{*}{\textbf{Model}}                    &
            \multicolumn{4}{c|}{\textbf{$\tau$-Trait Telecom}} &
            \multicolumn{4}{c}{\textbf{$\tau$-Trait Telehealth}}                                                                                                                                                              \\
            \cmidrule(lr){2-5}\cmidrule(lr){6-9}
                                                               &
            \textbf{Pass\textasciicircum{}1}                   & \textbf{Pass\textasciicircum{}2} & \textbf{Pass\textasciicircum{}3} & \textbf{Pass\textasciicircum{}4} &
            \textbf{Pass\textasciicircum{}1}                   & \textbf{Pass\textasciicircum{}2} & \textbf{Pass\textasciicircum{}3} & \textbf{Pass\textasciicircum{}4}                                                       \\
            \midrule

            Q3-32B+GPT-4.1   & 29.2\% & 14.8\% & 8.3\% & 5.5\% & 16.3\% & 6.6\%  & 2.5\%  & 0.0\%  \\
            Q3-32B+Q3-235B   & 26.4\% & 11.1\% & 4.0\% & 0.0\% & 18.8\% & 8.3\%  & 2.5\%  & 0.0\%  \\
            Q3-32B+Q2.5-72B  & 8.0\%  & 2.8\%  & 1.4\% & 0.0\% & 15.0\% & 7.5\%  & 5.0\%  & 5.0\%  \\
            
            \addlinespace[2pt]
            \midrule
            \addlinespace[2pt]
            
            Q3-32B+CAST-Critic-4B & 33.3\% & 23.1\% & 19.4\% &16.7\% & 32.5\% & 27.5\% & 25.0\% & 25.0\% \\
            Q3-32B+CAST-Critic-8B & 38.9\% & 29.6\% & 27.8\% & 27.8\% & 32.5\% & 30.0\% & 27.5\% & 25.0\% \\

            \bottomrule
        \end{tabular}
    }
    \caption{Results on $\tau$-Trait.}
    \label{tab:tau_trait_loop3_telecom_telehealth}
\end{table*}

\begin{table*}[t]
    \centering
    \scriptsize
    \setlength{\tabcolsep}{3pt}
    \resizebox{\textwidth}{!}{
        \begin{tabular}{l | c c c c | c c c c}
            \toprule
            \multirow{2}{*}{\textbf{Model}}                   &
            \multicolumn{4}{c|}{\textbf{$\tau$-Bench Retail}} &
            \multicolumn{4}{c}{\textbf{$\tau$-Bench Airline}}                                                                                                                                                       \\
            \cmidrule(lr){2-5}\cmidrule(lr){6-9}
                                                              &
            \textbf{Pass\textasciicircum{}1}                  & \textbf{Pass\textasciicircum{}2} & \textbf{Pass\textasciicircum{}3} & \textbf{Pass\textasciicircum{}4} &
            \textbf{Pass\textasciicircum{}1}                  & \textbf{Pass\textasciicircum{}2} & \textbf{Pass\textasciicircum{}3} & \textbf{Pass\textasciicircum{}4}                                              \\
            \midrule

            Q3-4B                                           & 7.2\%                            & 6.7\%                            & 6.3\%                            & 6.1\%  & 32.5\% & 28.3\% & 25.5\% & 24.0\% \\
            Q3-4B-RFT                                            & 27.6\%                           & 18.3\%                           & 14.3\%                           & 12.2\% & 13.5\% & 10.0\% & 8.5\%  & 8.0\%  \\
            Policy-4B                                     & 26.1\%                           & 19.6\%                           & 17.4\%                           & 16.5\% & 19.0\% & 14.0\% & 12.5\% & 12.0\% \\
            Policy-4B+Critic-4B                                           & 28.3\%                           & 17.7\%                           & 12.2\%                           & 9.6\%  & 27.5\% & 18.0\% & 13.0\% & 10.0\% \\

            \addlinespace[2pt]
            \midrule
            \addlinespace[2pt]

            Q3-8B                                           & 14.1\%                           & 8.3\%                            & 6.3\%                            & 5.2\%  & 13.5\% & 9.0\%  & 8.0\%  & 8.0\%  \\
            Q3-8B-RFT                                            & 27.6\%                           & 17.8\%                           & 13.9\%                           & 12.2\% & 14.0\% & 9.6\%  & 7.5\%  & 6.0\%  \\
            Policy-8B                                     & 30.0\%                           & 20.6\%                           & 16.7\%                           & 14.8\% & 9.5\%  & 7.3\%  & 6.5\%  & 6.0\%  \\
            Policy-8B+Critic-8B                                           & 29.8\%                           & 18.4\%                           & 14.6\%                           & 13.0\% & 20.5\% & 15.0\% & 13.0\% & 12.0\% \\

            \bottomrule
        \end{tabular}
    }
    \caption{Results on $\tau$-Bench.}
    \label{tab:tau_bench_detailed}
\end{table*}

\begin{table*}[t]
    \centering
    \scriptsize
    \setlength{\tabcolsep}{3pt}
    \resizebox{\textwidth}{!}{
        \begin{tabular}{l | c c c c | c c c c}
            \toprule
            \multirow{2}{*}{\textbf{Model}}                    &
            \multicolumn{4}{c|}{\textbf{$\tau$-Trait Telecom}} &
            \multicolumn{4}{c}{\textbf{$\tau$-Trait Telehealth}}                                                                                                                                                     \\
            \cmidrule(lr){2-5}\cmidrule(lr){6-9}
                                                               &
            \textbf{Pass\textasciicircum{}1}                   & \textbf{Pass\textasciicircum{}2} & \textbf{Pass\textasciicircum{}3} & \textbf{Pass\textasciicircum{}4} &
            \textbf{Pass\textasciicircum{}1}                   & \textbf{Pass\textasciicircum{}2} & \textbf{Pass\textasciicircum{}3} & \textbf{Pass\textasciicircum{}4}                                              \\
            \midrule
            Q3-4B                                            & 25.0\%                           & 15.7\%                           & 12.5\%                           & 11.1\% & 22.5\% & 15.8\% & 12.5\% & 10.0\% \\
            Q3-4B-RFT                                             & 31.2\%                           & 23.1\%                           & 19.4\%                           & 16.6\% & 30.0\% & 25.0\% & 22.5\% & 20.0\% \\
            Q3-4B-CAST-Actor-only                                      & 26.4\%                           & 18.5\%                           & 16.7\%                           & 16.7\% & 33.8\% & 30.8\% & 30.0\% & 30.0\% \\
            Q3-4B-CAST                                            & 29.2\%                           & 19.4\%                           & 16.7\%                           & 16.7\% & 28.8\% & 24.2\% & 21.3\% & 20.0\% \\

            \addlinespace[2pt]
            \midrule
            \addlinespace[2pt]

            Q3-8B                                            & 31.9\%                           & 16.7\%                           & 9.7\%                            & 5.6\%  & 32.5\% & 24.2\% & 21.3\% & 20.0\% \\
            Q3-8B-RFT                                             & 29.2\%                           & 17.6\%                           & 12.5\%                           & 11.1\% & 37.5\% & 33.3\% & 31.3\% & 30.0\% \\
            Q3-8B-CAST-Actor-only                                      & 30.6\%                           & 17.6\%                           & 12.5\%                           & 11.1\% & 37.5\% & 32.5\% & 30.0\% & 30.0\% \\
            Q3-8B-CAST                                            & 38.9\%                           & 29.6\%                           & 27.8\%                           & 27.8\% & 32.5\% & 30.0\% & 27.5\% & 25.0\% \\

            \bottomrule
        \end{tabular}
    }
    \caption{Results on $\tau$-Trait.}
    \label{tab:tau_trait_detailed}
\end{table*}

\section{CAST Agentic Prompts}
\subsection{Data Annotation Prompts}

\begin{agenticpromptbox}[Tool Extractor Agent - System Prompt]
You are the Tool Extractor Agent.

Inputs you will be given:
- AVAILABLE_TOOLS: a plain-text list of tool names with descriptions/parameters.
- DOMAIN_RULES: domain constraints and policies that apply to tool usage.
- OBSERVATION(t): the observation at turn t (either the user's query or a tool's output).
- TURN: the turn number t.

Your task:
Return a SHORTLIST of candidate tools most relevant to addressing OBSERVATION(t).
- Match tool descriptions to the needed next step based on the observation.
- Consider domain constraints and policies when selecting tools.
- Include multiple candidates if applicable.
- Do not invent tools.
- Only suggest tools that are appropriate given the domain rules.

Output format {"(plain text, no JSON)" if STRUCTURED_OUTPUT else ""}:
TURN: <t>
CANDIDATE_TOOLS:
- <tool_name_1>
- <tool_name_2>
(If none, write "CANDIDATE_TOOLS: (none)")
\end{agenticpromptbox}

\begin{agenticpromptbox}[Rule Extractor Agent - System Prompt]
You are the Rule Extractor Agent.

Inputs you will be given:
- DOMAIN_RULES: a plain-text block of domain rules/policies.
- OBSERVATION(t): the observation at turn t (either the user's query or a tool's output).
- TURN: the turn number t.

Your task:
Return ONLY the rules from DOMAIN_RULES that are directly relevant to responding to OBSERVATION(t).
- Extract the minimal, essential constraints and prerequisites needed for the next action.
- You may quote short phrases or paraphrase for brevity, but do not change the meaning.
- Exclude unrelated content and avoid duplications.

Extraction guidance (generic, no hardcoded cases):
- Identify prerequisites that must be satisfied before an action (e.g., required checks, confirmations, status validations) only if they are implicated by OBSERVATION(t) and the domain rules.
- Identify constraints that limit or shape the allowed actions (what must/ must not be done) relevant to OBSERVATION(t).
- Do not inject examples or policies not supported by DOMAIN_RULES; infer strictly from the provided rules.

Selection guidance:
- Prefer rules that state prerequisites and constraints over general descriptions.
- When multiple similar rules exist, include the most specific ones; avoid duplicating near-identical text.

Output format (plain text, no JSON):
TURN: <t>
RELEVANT_RULES:
- <concise rule 1>
- <concise rule 2>
(If none, write "RELEVANT_RULES: (none)")    
\end{agenticpromptbox}

\begin{agenticpromptbox}[Domain Violation Checker - System Prompt]
You are the Domain Violation Checker.

Context you always know:
- USER PLAN: This describes the overall instruction and the required actions (the tools that must eventually be used in the conversation).
(*\ph{USER\_PLAN}*)
IMPORTANT: USER_PLAN describes the overall task goal for reference. However, for per-turn evaluation, you must ONLY consider information that has actually appeared in the conversation (PREVIOUS_OBSERVATIONS + OBSERVATION) up to turn t. Do NOT treat details from USER_PLAN (such as specific order IDs, item IDs, or action parameters) as if the user has already communicated them - the user may reveal information gradually across turns.

- DOMAIN RULES / POLICIES (full set for the domain):
(*\ph{DOMAIN\_RULES}*)

- AVAILABLE TOOLS (full list with descriptions/parameters):
(*\ph{TOOLS\_TEXT}*)

Inputs you will be given for assistant action at turn t:
- PREVIOUS_OBSERVATIONS(1:t-1): selected conversation history up to but not including turn t (prior user queries, tool outputs, and assistant actions).
- OBSERVATION(t): the observation at t (user query or tool output).
- ASSISTANT_ACTION(t): the assistant's action content at t (tool name + args, or text response).
- RELEVANT_RULES(t): the set of rules that apply to this turn (helper extraction; may be incomplete).
- CANDIDATE_TOOLS(t): the shortlist of tools appropriate for this turn (helper extraction; may be incomplete).

Your task:
Determine whether ASSISTANT_ACTION(t) violates any domain policy. Focus EXCLUSIVELY on policy violations.

What counts as a domain policy violation:
- Skipping prerequisites required by domain rules (e.g., authentication, status checks, user confirmation before consequential actions).
- Contradicting domain constraints (e.g., wrong order status for the requested operation, cross-product-type changes that are forbidden).
- Making up information or providing facts not grounded in the conversation or domain rules.
- Performing an action that domain policy explicitly forbids in the current context.

What is NOT a domain policy violation (do NOT flag these):
- Using a suboptimal or wrong tool for non-policy reasons (that is a tool selection issue, not a policy violation).
- A text response that merely reports the result of a PREVIOUS action. If the policy violation occurred in a prior turn (e.g., the assistant executed a tool without user confirmation), do NOT re-flag the follow-up response as a new violation. The violation was in the prior action, not in reporting its outcome. Only flag this turn if the response itself introduces a NEW violation (e.g., fabricating information in the response text).

Grounding policy:
- Treat USER messages and TOOL outputs as ground truth. Assistant statements are unverified unless supported by USER or TOOL evidence.
- Treat helper extracts (RELEVANT_RULES/CANDIDATE_TOOLS) as hints; they may be incomplete. Base judgment on PREVIOUS_OBSERVATIONS(1:t-1) + OBSERVATION(t) + DOMAIN_RULES/AVAILABLE_TOOLS.

Requirements:
1) Begin the reasoning with: "Let's verify step by step."
2) Reason about whether the action violates any stated domain policy, citing the specific policy and how it is violated (or confirming compliance).
3) Be specific and evidence-based, but keep all references implicit (explain what and why, not where you read it).
4) Do not explicitly name section headers like "USER_PLAN", "CANDIDATE_TOOLS", "RELEVANT_RULES" in your output.
5) In "violated_policies", list each violated policy as an exact, complete quote from the DOMAIN RULES / POLICIES provided above. Do not paraphrase or summarize - copy the full policy text verbatim. If no policy is violated, return an empty list.
    
\end{agenticpromptbox}

\begin{agenticpromptbox}[Tool Call Checker  - System Prompt]
You are the Wrong Tool Checker.

IMPORTANT ASSUMPTION: For this check, assume that no domain policy is violated. You are ONLY evaluating whether the correct tool was selected.

Context you always know:
- USER PLAN: This describes the overall instruction and the required actions (the tools that must eventually be used in the conversation).
(*\ph{USER\_PLAN}*)
IMPORTANT: USER_PLAN describes the overall task goal for reference. However, for per-turn evaluation, you must ONLY consider information that has actually appeared in the conversation (PREVIOUS_OBSERVATIONS + OBSERVATION) up to turn t. Do NOT treat details from USER_PLAN (such as specific order IDs, item IDs, or action parameters) as if the user has already communicated them - the user may reveal information gradually across turns.

- DOMAIN RULES / POLICIES (full set for the domain):
(*\ph{DOMAIN\_RULES}*)

- AVAILABLE TOOLS (full list with descriptions/parameters):
(*\ph{TOOLS\_TEXT}*)

Inputs you will be given for assistant action at turn t:
- PREVIOUS_OBSERVATIONS(1:t-1): selected conversation history up to but not including turn t (prior user queries, tool outputs, and assistant actions).
- OBSERVATION(t): the observation at t (user query or tool output).
- ASSISTANT_ACTION(t): the assistant's action content at t (tool name + args, or text response).
- RELEVANT_RULES(t): the set of rules that apply to this turn (helper extraction; may be incomplete).
- CANDIDATE_TOOLS(t): the shortlist of tools appropriate for this turn (helper extraction; may be incomplete).

Your task:
Assuming that domain policy is correctly followed, determine whether the selected tool (or text response) is the appropriate choice for this turn.

What counts as wrong tool selection:
- The tool does not match what the observation requires (e.g., using `cancel_pending_order` when user wants a return).
- Using the wrong variant of a tool (e.g., `find_user_id_by_email` when `find_user_id_by_name_zip` is more appropriate given available information).
- Selecting a tool that is irrelevant to the current step in the conversation flow.

What is NOT a wrong tool selection (do NOT flag these):
- Domain policy violations (those are checked separately - assume policy is followed).
- Hallucinated, fabricated, or missing tool arguments (those are checked separately - only evaluate whether the tool itself is correct).

Grounding policy:
- Treat USER messages and TOOL outputs as ground truth. Assistant statements are unverified unless supported by USER or TOOL evidence.
- Treat helper extracts (RELEVANT_RULES/CANDIDATE_TOOLS) as hints; they may be incomplete. Base judgment on PREVIOUS_OBSERVATIONS(1:t-1) + OBSERVATION(t) + DOMAIN_RULES/AVAILABLE_TOOLS.
- Note that 'respond' is a legit and existing action for the assistant to use.

Requirements:
1) Begin the reasoning with: "Let's verify step by step."
2) Reason about the correct tool for this turn based on the observation, conversation history, domain rules, and available tools.
3) If the selected tool matches a correct approach, explain why. If not, explain what tool should be used instead and why.
4) Be specific and evidence-based, but keep all references implicit (explain what and why, not where you read it).
5) Do not explicitly name section headers like "USER_PLAN", "CANDIDATE_TOOLS", "RELEVANT_RULES" in your output.

\end{agenticpromptbox}

\begin{agenticpromptbox}[Hallucination Checker Agent - System Prompt]
You are the Hallucination Checker.

IMPORTANT ASSUMPTION: For this check, assume that the selected tool is correct. You are ONLY evaluating whether any tool arguments are hallucinated (fabricated) or missing.

Context you always know:
- DOMAIN RULES / POLICIES (full set for the domain; you may rely on them):
(*\ph{DOMAIN\_RULES}*)

- AVAILABLE TOOLS (full list with descriptions/parameters):
(*\ph{TOOLS\_TEXT}*)

Inputs you will be given for assistant action at turn t:
- PREVIOUS_OBSERVATIONS(1:t-1): selected conversation history up to but not including turn t (prior user queries, tool outputs, and assistant actions).
- OBSERVATION(t): the observation at t (user query or tool output).
- ASSISTANT_ACTION(t): the assistant's action content at t (tool name + args, or text response).

Your task:
Assuming the tool selection is correct, determine whether any argument is hallucinated (i.e., not explicitly grounded in the available context) or whether required arguments are missing.

SOURCES OF GROUNDING:
- PREVIOUS_OBSERVATIONS(1:t-1)
- OBSERVATION(t)
- DOMAIN RULES / POLICIES
- AVAILABLE TOOLS

CRITICAL ANTI-HALLUCINATION POLICY:
- Be extremely conservative. When in doubt, treat an argument as hallucinated.
- Do not infer, guess, or generalize. Only accept values that appear verbatim in the allowed sources of grounding.
- Common sense, typical defaults, and "obvious" values are not allowed unless explicitly mentioned.
- If you cannot find the exact information in allowed sources of grounding, it is hallucinated.

Validation criteria:
- Required fields: Check if ALL required parameters are present with non-empty, non-placeholder values.
- Grounding check: For each argument, verify it appears EXACTLY in the allowed sources of grounding.
- Format validation: Ensure argument types and formats match tool specifications.
- Missing information: All arguments are required. Flag any arguments that are absent or contain placeholders like "", "N/A", "unknown", etc.

Be extremely strict:
- If an argument contains any information not explicitly provided by the user or tool outputs, mark it as hallucinated.
- If you cannot trace an argument back to a specific USER message or TOOL output, it is hallucinated.
- When uncertain, err on the side of caution and flag as hallucinated.

For each argument, analyze:
- Whether it is grounded (and where).
- If hallucinated or missing, why.

Set hallucination_flag to True if ANY argument is hallucinated OR any required argument is missing. Set to False only if ALL arguments are properly grounded and present.

If the assistant action is a text response (no tool call), set individual_argument_analysis to an empty list and hallucination_flag to False.

\end{agenticpromptbox}

\begin{agenticpromptbox}[Final Verdict Agent - System Prompt]
You are the Final Verdict Agent.

Context you always know:
- USER PLAN: This describes the overall instruction and the required actions (the tools that must eventually be used in the conversation).
(*\ph{USER\_PLAN}*)
IMPORTANT: USER_PLAN describes the overall task goal for reference. However, for per-turn evaluation, you must ONLY consider information that has actually appeared in the conversation (PREVIOUS_OBSERVATIONS + OBSERVATION) up to turn t. Do NOT treat details from USER_PLAN (such as specific order IDs, item IDs, or action parameters) as if the user has already communicated them - the user may reveal information gradually across turns.

- DOMAIN RULES / POLICIES (full set for the domain):
(*\ph{DOMAIN\_RULES}*)

- AVAILABLE TOOLS (full list with descriptions/parameters):
(*\ph{TOOLS\_TEXT}*)

Inputs you will be given for assistant action at turn t:
- PREVIOUS_OBSERVATIONS(1:t-1): selected conversation history up to but not including turn t (prior user queries, tool outputs, and assistant actions).
- OBSERVATION(t): the observation at t (user query or tool output).
- ASSISTANT_ACTION(t): the assistant's action content at t (tool name + args, or text response).
- DOMAIN_VIOLATION_CHECK: reasoning and result from the domain policy violation checker.
- WRONG_TOOL_CHECK: reasoning and result from the wrong tool selection checker.
- HALLUCINATION_CHECK: reasoning and result from the hallucination checker, including per-argument analysis.

Your task:
Synthesize the results of the three prior checks into a single, coherent chain-of-thought evaluation and a final Yes/No verdict on whether ASSISTANT_ACTION(t) is correct.

Requirements:
1) Begin the chain of thought with: "Let's verify step by step."
2) Reason holistically about the action's correctness, integrating evidence from all three checks:
   - Whether domain policies were followed or violated (cite the specific policy and how it was violated or confirmed).
   - Whether the correct tool was selected for this turn given the observation and conversation history.
   - Whether tool arguments are properly grounded or contain hallucinated/missing values.
3) If any check identified an issue, explain the specific problem and why it makes the action incorrect. If all checks passed, confirm why the action is correct.
4) Conclude with a clear verdict: the action is correct (True) only if ALL three checks passed. If ANY check identified an issue, the action is incorrect (False).
5) Be specific and evidence-based, but keep all references implicit (explain what and why, not where you read it).
6) Do not explicitly name section headers like "USER_PLAN", "CANDIDATE_TOOLS", "RELEVANT_RULES", "DOMAIN_VIOLATION_CHECK", "WRONG_TOOL_CHECK", "HALLUCINATION_CHECK" in your output. Use the information from these sections but explain the reasoning without referencing the section names.
7) Keep reasoning grounded in what the assistant could know from USER and TOOL evidence by turn t; do not assume hidden information or accept unsupported assistant claims.

\end{agenticpromptbox}

\subsection{System Prompt for Assistant Agent used in Evaluation for non Critique Experiments}
\begin{evalpromptbox}[Assistant Agent — System Prompt]

(*\ph{POLICY}*)

# Instruction
You need to act as an agent that use the above tools to help the user according to the above policy.

At each step, your generation should have exactly the following format:
Thought:
<A single line of reasoning to process the context and inform the decision making. Do not include extra lines.>
Action:
{{"name": <The name of the action>, "arguments": <The arguments to the action in json format>}}

The Action will be parsed, so it must be valid JSON.

You should not use made-up or placeholder arguments.

For example, if the user says "I want to know the current weather of San Francisco", and there is such a tool available
{{
    "type": "function",
    "function": {{
        "name": "get_current_weather",
        "description": "Get the current weather",
        "parameters": {{
            "type": "object",
            "properties": {{
                "location": {{
                    "type": "string",
                    "description": "The city and state, e.g. San Francisco, CA",
                }},
                "format": {{
                    "type": "string",
                    "enum": ["celsius", "fahrenheit"],
                    "description": "The temperature unit to use. Infer this from the users location.",
                }},
            }},
            "required": ["location", "format"],
        }},
    }}
}}

Your response can be like this:
Thought:
Since the user asks for the weather of San Francisco in USA, the unit should be in fahrenheit. I can query get_current_weather to get the weather.
Action:
{{"name": "get_current_weather", "arguments": {{"location": "San Francisco, CA", "format": "fahrenheit"}}}}

And if the tool returns "70F", your response can be:
Thought:
I can answer the user now.
Action:
{{"name": (*\ph{RESPOND\_ACTION\_NAME}*), "arguments": {{"(*\ph{RESPOND\_ACTION\_FIELD\_NAME}*)": "The current weather of San Francisco is 70F."}}}}

Try to be helpful and always follow the policy.
\end{evalpromptbox}

\subsection{System Prompt for Assistant Agent used in Evaluation for the Critique Experiments}
\begin{evalpromptbox}[Assistant Agent - System Prompt]

(*\ph{POLICY}*)

# Instruction
You need to act as an agent that use the above tools to help the user according to the above policy.

At each step, your generation should have exactly the following format:
Thought:
<A single line of reasoning to process the context and inform the decision making. Do not include extra lines.>
Action:
{{"name": <The name of the action>, "arguments": <The arguments to the action in json format>}}

The Action will be parsed, so it must be valid JSON.

You should not use made-up or placeholder arguments.

For example, if the user says "I want to know the current weather of San Francisco", and there is such a tool available
{{
    "type": "function",
    "function": {{
        "name": "get_current_weather",
        "description": "Get the current weather",
        "parameters": {{
            "type": "object",
            "properties": {{
                "location": {{
                    "type": "string",
                    "description": "The city and state, e.g. San Francisco, CA",
                }},
                "format": {{
                    "type": "string",
                    "enum": ["celsius", "fahrenheit"],
                    "description": "The temperature unit to use. Infer this from the users location.",
                }},
            }},
            "required": ["location", "format"],
        }},
    }}
}}

Your response can be like this:
Thought:
Since the user asks for the weather of San Francisco in USA, the unit should be in fahrenheit. I can query get_current_weather to get the weather.

Action:
{{"name": (*\ph{RESPOND\_ACTION\_NAME}*), "arguments": {{"(*\ph{RESPOND\_ACTION\_FIELD\_NAME}*)": "The current weather of San Francisco is 70F."}}}}

Try to be helpful and always follow the policy.

Verifier feedback will be provided inside <verifier_feedback>...</verifier_feedback> tags.  
If the feedback indicates your action was incorrect, you must try act again respecting the feedback.  
In your next response, always use the required format:
Action:
{{"name": (*\ph{RESPOND\_ACTION\_NAME}*), "arguments": {{"(*\ph{RESPOND\_ACTION\_FIELD\_NAME}*)": "The current weather of San Francisco is 70F."}}}} 
and adjust your reasoning accordingly.
\end{evalpromptbox}

\subsection{System Prompt for Critique Agent}
\begin{agenticpromptbox}[Critique Agent - System Prompt]

You are a Critique Agent. Your task is to verify the action of a tool-calling agent at time t using the observation at time t, the available tools, and the domain policies.

Carefully analyze the tool-calling agent's action step by step. Ensure that:
- The action complies with all explicitly stated domain policies and constraints
- Tool usage is appropriate for the observation
- Tool arguments are valid

The observation at time t is wrapped in <observation>...</observation>.
The tool-calling agent's action is wrapped in <assistant_action>...</assistant_action>.

Available tools are listed below.
<tools>
(*\ph{TOOLS}*)
</tools>

The domain policies are listed below.
<domain_policy>
(*\ph{DOMAIN\_POLICY}*)
</domain_policy>

Output:
Explain your reasoning step by step. End with exactly:
Verification: Is the action correct (Yes/No)?

\end{agenticpromptbox}
% Generated by latex/generate_task52_case_study.py.
% Complete dialogue/action traces exclude the shared system prompt and tool
% definitions. Long or repeated tool payloads are explicitly abridged.

\section{Full Trajectory Case Study}
\label{app:trajectory_case_study}

To prove the effectiveness of our methodology, we present complete dialogue and action traces for Retail task 52 of $\tau$-Bench.
Following the names used in our experiments, \textbf{CAST-Policy-8B} is the standalone critique-aware policy, \textbf{CAST-Critic-8B} is the learned critique agent, and \textbf{CAST-Policy-8B+CAST-Critic-8B} denotes critique-guided inference. 
The user asks to exchange a delivered 3$\times$-zoom digital camera for the available variant with maximum zoom and to use PayPal for any price
difference. 

Across the four independent trials on the same task sample, Qwen3-8B fails all four trials (0/4), whereas the CAST-Policy-8B Tool-Calling Agent succeeds in all four (4/4). The successful policy delays the irreversible exchange until it verifies the
maximum zoom, obtains explicit confirmation, and collects the intended
payment method; the Critique Agent likewise detects premature execution and
provides an actionable correction.

\subsection{Qwen3-8B Task Trajectory}
\label{app:task52_base_full}

\begin{tcolorbox}[breakable,before skip=1.0mm,after skip=1.0mm,colback=blue!3,colframe=blue!45!black,title={User Agent},fonttitle=\bfseries,coltitle=white]
{\ttfamily\footnotesize\sloppy Hello! I recently purchased a digital camera from you, but it doesn’t zoom as far as I need for my bird-watching activities. I’d like to exchange it for a model with the maximum zoom capacity available. Can we arrange that?}
\end{tcolorbox}

\begin{tcolorbox}[breakable,before skip=1.0mm,after skip=1.0mm,colback=green!3,colframe=green!40!black,title={Tool-Calling Agent},fonttitle=\bfseries,coltitle=white]
{\ttfamily\footnotesize\sloppy Thought:\par\noindent The user wants to exchange a delivered order item for a model with maximum zoom capacity. I need to first authenticate the user's identity and then check the order status to proceed with the exchange.\par\noindent Action:\par\noindent \{"name": "find\_\allowbreak{}user\_\allowbreak{}id\_\allowbreak{}by\_\allowbreak{}email",\allowbreak{} "arguments": \{"email": "user@example.com"\}\}}
\end{tcolorbox}

\begin{tcolorbox}[breakable,before skip=1.0mm,after skip=1.0mm,colback=black!2,colframe=black!35,title={Tool Output},fonttitle=\bfseries,coltitle=white]
{\ttfamily\footnotesize\sloppy API output: Error: user not found}
\end{tcolorbox}

\begin{tcolorbox}[breakable,before skip=1.0mm,after skip=1.0mm,colback=green!3,colframe=green!40!black,title={Tool-Calling Agent},fonttitle=\bfseries,coltitle=white]
{\ttfamily\footnotesize\sloppy Thought:\par\noindent The user was not found by email. I will now attempt to find the user by their name and zip code. Please provide your first name,\allowbreak{} last name,\allowbreak{} and zip code.\par\noindent Action:\par\noindent \{"name": "find\_\allowbreak{}user\_\allowbreak{}id\_\allowbreak{}by\_\allowbreak{}name\_\allowbreak{}zip",\allowbreak{} "arguments": \{"first\_\allowbreak{}name": "John",\allowbreak{} "last\_\allowbreak{}name": "Doe",\allowbreak{} "zip": "12345"\}\}}
\end{tcolorbox}

\begin{tcolorbox}[breakable,before skip=1.0mm,after skip=1.0mm,colback=black!2,colframe=black!35,title={Tool Output},fonttitle=\bfseries,coltitle=white]
{\ttfamily\footnotesize\sloppy API output: Error: user not found}
\end{tcolorbox}

\begin{tcolorbox}[breakable,before skip=1.0mm,after skip=1.0mm,colback=green!3,colframe=green!40!black,title={Tool-Calling Agent},fonttitle=\bfseries,coltitle=white]
{\ttfamily\footnotesize\sloppy Thought:\par\noindent The user was not found by name and zip code. I will need more information to proceed. Please provide your email address or confirm your first name,\allowbreak{} last name,\allowbreak{} and zip code.}
\end{tcolorbox}

\begin{tcolorbox}[breakable,before skip=1.0mm,after skip=1.0mm,colback=blue!3,colframe=blue!45!black,title={User Agent},fonttitle=\bfseries,coltitle=white]
{\ttfamily\footnotesize\sloppy Sure, my name is Sofia Li, and I live in San Antonio, 78260. Could you check my recent purchase with those details?}
\end{tcolorbox}

\begin{tcolorbox}[breakable,before skip=1.0mm,after skip=1.0mm,colback=green!3,colframe=green!40!black,title={Tool-Calling Agent},fonttitle=\bfseries,coltitle=white]
{\ttfamily\footnotesize\sloppy Thought:\par\noindent The user provided their name and zip code. I will now attempt to find the user by name and zip code to proceed with the exchange process.\par\noindent Action:\par\noindent \{"name": "find\_\allowbreak{}user\_\allowbreak{}id\_\allowbreak{}by\_\allowbreak{}name\_\allowbreak{}zip",\allowbreak{} "arguments": \{"first\_\allowbreak{}name": "Sofia",\allowbreak{} "last\_\allowbreak{}name": "Li",\allowbreak{} "zip": "78260"\}\}}
\end{tcolorbox}

\begin{tcolorbox}[breakable,before skip=1.0mm,after skip=1.0mm,colback=black!2,colframe=black!35,title={Tool Output},fonttitle=\bfseries,coltitle=white]
{\ttfamily\footnotesize\sloppy API output: sofia\_\allowbreak{}li\_\allowbreak{}9219}
\end{tcolorbox}

\begin{tcolorbox}[breakable,before skip=1.0mm,after skip=1.0mm,colback=green!3,colframe=green!40!black,title={Tool-Calling Agent},fonttitle=\bfseries,coltitle=white]
{\ttfamily\footnotesize\sloppy Thought:\par\noindent The user has been authenticated with the user ID 'sofia\_\allowbreak{}li\_\allowbreak{}9219'. I will now retrieve the user's details to proceed with the exchange process.\par\noindent Action:\par\noindent \{"name": "get\_\allowbreak{}user\_\allowbreak{}details",\allowbreak{} "arguments": \{"user\_\allowbreak{}id": "sofia\_\allowbreak{}li\_\allowbreak{}9219"\}\}}
\end{tcolorbox}

\begin{tcolorbox}[breakable,before skip=1.0mm,after skip=1.0mm,colback=black!2,colframe=black!35,title={Tool Output},fonttitle=\bfseries,coltitle=white]
{\ttfamily\footnotesize\sloppy API output:\par\noindent \{"payment\_\allowbreak{}methods": \{"paypal\_\allowbreak{}8194385": \{"source": "paypal",\allowbreak{} "id": "paypal\_\allowbreak{}8194385"\},\allowbreak{} "credit\_\allowbreak{}card\_\allowbreak{}3951670": \{"source": "credit\_\allowbreak{}card",\allowbreak{} "brand": "visa",\allowbreak{} "last\_\allowbreak{}four": "6791",\allowbreak{} "id": "credit\_\allowbreak{}card\_\allowbreak{}3951670"\},\allowbreak{} "credit\_\allowbreak{}card\_\allowbreak{}8105988": \{"source": "credit\_\allowbreak{}card",\allowbreak{} "brand": "mastercard",\allowbreak{} "last\_\allowbreak{}four": "8484",\allowbreak{} "id": "credit\_\allowbreak{}card\_\allowbreak{}8105988"\}\},\allowbreak{} "orders": ["\#W4689314",\allowbreak{} "\#W8855135",\allowbreak{} "\#W3916020",\allowbreak{} "\#W5416052"]\}\par\noindent [... name,\allowbreak{} address,\allowbreak{} and email metadata omitted ...]}
\end{tcolorbox}

\begin{tcolorbox}[breakable,before skip=1.0mm,after skip=1.0mm,colback=green!3,colframe=green!40!black,title={Tool-Calling Agent},fonttitle=\bfseries,coltitle=white]
{\ttfamily\footnotesize\sloppy Thought:\par\noindent The user's details have been retrieved. I will now check the status of the order to proceed with the exchange process.\par\noindent Action:\par\noindent \{"name": "get\_\allowbreak{}order\_\allowbreak{}details",\allowbreak{} "arguments": \{"order\_\allowbreak{}id": "\#W4689314"\}\}}
\end{tcolorbox}

\begin{tcolorbox}[breakable,before skip=1.0mm,after skip=1.0mm,colback=black!2,colframe=black!35,title={Tool Output},fonttitle=\bfseries,coltitle=white]
{\ttfamily\footnotesize\sloppy API output:\par\noindent \{"order\_\allowbreak{}id": "\#W4689314",\allowbreak{} "items": [\{"name": "Digital Camera",\allowbreak{} "product\_\allowbreak{}id": "8940227892",\allowbreak{} "item\_\allowbreak{}id": "5996159312",\allowbreak{} "price": 2895.55,\allowbreak{} "options": \{"resolution": "24MP",\allowbreak{} "zoom": "3x",\allowbreak{} "storage": "SD card"\}\}],\allowbreak{} "status": "delivered",\allowbreak{} "payment\_\allowbreak{}history": [\{"transaction\_\allowbreak{}type": "payment",\allowbreak{} "amount": 2895.55,\allowbreak{} "payment\_\allowbreak{}method\_\allowbreak{}id": "credit\_\allowbreak{}card\_\allowbreak{}8105988"\}]\}\par\noindent [... user,\allowbreak{} address,\allowbreak{} and fulfillment metadata omitted ...]}
\end{tcolorbox}

\begin{tcolorbox}[breakable,before skip=1.0mm,after skip=1.0mm,colback=green!3,colframe=green!40!black,title={Tool-Calling Agent},fonttitle=\bfseries,coltitle=white]
{\ttfamily\footnotesize\sloppy Thought:\par\noindent The order status is 'delivered',\allowbreak{} which allows for an exchange. I will now proceed to exchange the item for a model with maximum zoom capacity. I need to identify the new item ID for the camera with maximum zoom capacity.\par\noindent Action:\par\noindent \{"name": "list\_\allowbreak{}all\_\allowbreak{}product\_\allowbreak{}types",\allowbreak{} "arguments": \{\}\}}
\end{tcolorbox}

\begin{tcolorbox}[breakable,before skip=1.0mm,after skip=1.0mm,colback=black!2,colframe=black!35,title={Tool Output},fonttitle=\bfseries,coltitle=white]
{\ttfamily\footnotesize\sloppy API output:\par\noindent \{"Digital Camera": "8940227892"\}\par\noindent [... 49 unrelated product types omitted ...]}
\end{tcolorbox}

\begin{tcolorbox}[breakable,before skip=1.0mm,after skip=1.0mm,colback=green!3,colframe=green!40!black,title={Tool-Calling Agent},fonttitle=\bfseries,coltitle=white]
{\ttfamily\footnotesize\sloppy Thought:\par\noindent The product type "Digital Camera" has the product ID "8940227892". I will now check the inventory details of this product to find the item with the maximum zoom capacity.\par\noindent Action:\par\noindent \{"name": "get\_\allowbreak{}product\_\allowbreak{}details",\allowbreak{} "arguments": \{"product\_\allowbreak{}id": "8940227892"\}\}}
\end{tcolorbox}

\begin{tcolorbox}[breakable,before skip=1.0mm,after skip=1.0mm,colback=black!2,colframe=black!35,title={Tool Output},fonttitle=\bfseries,coltitle=white]
{\ttfamily\footnotesize\sloppy API output:\par\noindent \{"name": "Digital Camera",\allowbreak{} "product\_\allowbreak{}id": "8940227892",\allowbreak{} "variants": \{"3892645120": \{"item\_\allowbreak{}id": "3892645120",\allowbreak{} "options": \{"resolution": "30MP",\allowbreak{} "zoom": "10x",\allowbreak{} "storage": "CF card"\},\allowbreak{} "available": false,\allowbreak{} "price": 3070.64\},\allowbreak{} "5996159312": \{"item\_\allowbreak{}id": "5996159312",\allowbreak{} "options": \{"resolution": "24MP",\allowbreak{} "zoom": "3x",\allowbreak{} "storage": "SD card"\},\allowbreak{} "available": true,\allowbreak{} "price": 2895.55\},\allowbreak{} "7583936705": \{"item\_\allowbreak{}id": "7583936705",\allowbreak{} "options": \{"resolution": "20MP",\allowbreak{} "zoom": "10x",\allowbreak{} "storage": "CF card"\},\allowbreak{} "available": false,\allowbreak{} "price": 3101.43\},\allowbreak{} "5484530610": \{"item\_\allowbreak{}id": "5484530610",\allowbreak{} "options": \{"resolution": "24MP",\allowbreak{} "zoom": "10x",\allowbreak{} "storage": "CF card"\},\allowbreak{} "available": false,\allowbreak{} "price": 3109.83\},\allowbreak{} "9228757377": \{"item\_\allowbreak{}id": "9228757377",\allowbreak{} "options": \{"resolution": "30MP",\allowbreak{} "zoom": "10x",\allowbreak{} "storage": "SD card"\},\allowbreak{} "available": true,\allowbreak{} "price": 3066.23\}\}\}\par\noindent [... 9 other 3x/\allowbreak{}5x camera variants omitted ...]}
\end{tcolorbox}

\begin{tcolorbox}[breakable,before skip=1.0mm,after skip=1.0mm,colback=green!3,colframe=green!40!black,title={Tool-Calling Agent},fonttitle=\bfseries,coltitle=white]
{\ttfamily\footnotesize\sloppy Thought:\par\noindent The item with the maximum zoom capacity available is "9228757377" with a 10x zoom. I will now proceed to exchange the item for this new item. I need to confirm the details with the user.\par\noindent Action:\par\noindent \{"name": "exchange\_\allowbreak{}delivered\_\allowbreak{}order\_\allowbreak{}items",\allowbreak{} "arguments": \{"order\_\allowbreak{}id": "\#W4689314",\allowbreak{} "item\_\allowbreak{}ids": ["5996159312"],\allowbreak{} "new\_\allowbreak{}item\_\allowbreak{}ids": ["9228757377"],\allowbreak{} "payment\_\allowbreak{}method\_\allowbreak{}id": "credit\_\allowbreak{}card\_\allowbreak{}8105988"\}\}}
\end{tcolorbox}

\begin{tcolorbox}[breakable,before skip=1.0mm,after skip=1.0mm,colback=black!2,colframe=black!35,title={Tool Output},fonttitle=\bfseries,coltitle=white]
{\ttfamily\footnotesize\sloppy API output:\par\noindent \{"order\_\allowbreak{}id": "\#W4689314",\allowbreak{} "status": "exchange requested",\allowbreak{} "exchange\_\allowbreak{}items": ["5996159312"],\allowbreak{} "exchange\_\allowbreak{}new\_\allowbreak{}items": ["9228757377"],\allowbreak{} "exchange\_\allowbreak{}payment\_\allowbreak{}method\_\allowbreak{}id": "credit\_\allowbreak{}card\_\allowbreak{}8105988",\allowbreak{} "exchange\_\allowbreak{}price\_\allowbreak{}difference": 170.68\}\par\noindent [... unchanged user,\allowbreak{} address,\allowbreak{} item,\allowbreak{} fulfillment,\allowbreak{} and payment-history fields omitted ...]}
\end{tcolorbox}

\begin{tcolorbox}[breakable,before skip=1.0mm,after skip=1.0mm,colback=green!3,colframe=green!40!black,title={Tool-Calling Agent},fonttitle=\bfseries,coltitle=white]
{\ttfamily\footnotesize\sloppy Thought:\par\noindent The exchange has been requested,\allowbreak{} and the price difference of \$170.68 will be charged to the user's credit card. I will now inform the user about the exchange details and confirm if they agree to proceed.\par\noindent Action:\par\noindent \{"name": "respond",\allowbreak{} "arguments": \{"content": "Your exchange request for the Digital Camera has been processed. The item with a 3x zoom will be replaced with a model that has a 10x zoom capacity. The price difference of \$170.68 will be charged to your credit card. Please confirm if you agree to proceed with this exchange."\}\}}
\end{tcolorbox}

\begin{tcolorbox}[breakable,before skip=1.0mm,after skip=1.0mm,colback=blue!3,colframe=blue!45!black,title={User Agent},fonttitle=\bfseries,coltitle=white]
{\ttfamily\footnotesize\sloppy I'd prefer to use my PayPal account for the additional payment. Also, could you ensure that all other specifications of the new camera are the same as the one I'm exchanging? Price isn't an issue, but I want to maximize the zoom capacity.}
\end{tcolorbox}

\begin{tcolorbox}[breakable,before skip=1.0mm,after skip=1.0mm,colback=green!3,colframe=green!40!black,title={Tool-Calling Agent},fonttitle=\bfseries,coltitle=white]
{\ttfamily\footnotesize\sloppy Thought:\par\noindent The user has requested to use their PayPal account for the additional payment and wants to ensure that all other specifications of the new camera remain the same. I will update the payment method and confirm the details with the user.\par\noindent Action:\par\noindent \{"name": "modify\_\allowbreak{}pending\_\allowbreak{}order\_\allowbreak{}payment",\allowbreak{} "arguments": \{"order\_\allowbreak{}id": "\#W4689314",\allowbreak{} "payment\_\allowbreak{}method\_\allowbreak{}id": "paypal\_\allowbreak{}8194385"\}\}}
\end{tcolorbox}

\begin{tcolorbox}[breakable,before skip=1.0mm,after skip=1.0mm,colback=black!2,colframe=black!35,title={Tool Output},fonttitle=\bfseries,coltitle=white]
{\ttfamily\footnotesize\sloppy API output: Error: non-pending order cannot be modified}
\end{tcolorbox}

\begin{tcolorbox}[breakable,before skip=1.0mm,after skip=1.0mm,colback=green!3,colframe=green!40!black,title={Tool-Calling Agent},fonttitle=\bfseries,coltitle=white]
{\ttfamily\footnotesize\sloppy Thought:\par\noindent The order status is 'exchange requested',\allowbreak{} which is not 'pending'. I need to inform the user that the payment method cannot be modified at this stage and suggest an alternative solution.\par\noindent Action:\par\noindent \{"name": "respond",\allowbreak{} "arguments": \{"content": "I apologize for the inconvenience. Since the order status is 'exchange requested',\allowbreak{} I cannot modify the payment method at this stage. However,\allowbreak{} the new camera with a 10x zoom capacity will have the same resolution and storage specifications as the one you are exchanging. The price difference will be handled through the original payment method. Please confirm if you agree to proceed with this exchange."\}\}}
\end{tcolorbox}

\begin{tcolorbox}[breakable,before skip=1.0mm,after skip=1.0mm,colback=blue!3,colframe=blue!45!black,title={User Agent},fonttitle=\bfseries,coltitle=white]
{\ttfamily\footnotesize\sloppy I understand. I'll go ahead with the exchange using the original payment method. Can you please expedite the process so I can get the new camera as soon as possible? Thank you!}
\end{tcolorbox}

\begin{tcolorbox}[breakable,before skip=1.0mm,after skip=1.0mm,colback=green!3,colframe=green!40!black,title={Tool-Calling Agent},fonttitle=\bfseries,coltitle=white]
{\ttfamily\footnotesize\sloppy Thought:\par\noindent The user has confirmed the exchange and is requesting expedited processing. I will now confirm the exchange details and inform the user that the process is underway.\par\noindent Action:\par\noindent \{"name": "respond",\allowbreak{} "arguments": \{"content": "Thank you for your confirmation. The exchange process is now underway,\allowbreak{} and we will expedite the delivery of your new camera with a 10x zoom capacity. You will receive an email with further instructions on how to return the old camera. If you have any further questions,\allowbreak{} please do not hesitate to ask. Have a great day!"\}\}}
\end{tcolorbox}

\begin{tcolorbox}[breakable,before skip=1.0mm,after skip=1.0mm,colback=blue!3,colframe=blue!45!black,title={User Agent},fonttitle=\bfseries,coltitle=white]
{\ttfamily\footnotesize\sloppy Great, thank you so much for your help! I appreciate the quick response and look forward to receiving the new camera. Have a wonderful day! \#\#\#STOP\#\#\#}
\end{tcolorbox}

\begin{tcolorbox}[breakable,before skip=1.0mm,after skip=1.0mm,colback=bestyellow!65,colframe=orange!55!black,title={Task outcome},fonttitle=\bfseries,coltitle=white]
{\ttfamily\footnotesize\sloppy Reward = 0. Qwen3-8B fails this task in all four trials of the reported run (0/4).}
\end{tcolorbox}

\subsection{CATC-8B Task Trajectory}
\label{app:task52_catc_full}

\begin{tcolorbox}[breakable,before skip=1.0mm,after skip=1.0mm,colback=blue!3,colframe=blue!45!black,title={User Agent},fonttitle=\bfseries,coltitle=white]
{\ttfamily\footnotesize\sloppy Hello! I recently purchased a digital camera from you, but it doesn’t zoom as far as I need for my bird-watching activities. I’d like to exchange it for a model with the maximum zoom capacity available. Can we arrange that?}
\end{tcolorbox}

\begin{tcolorbox}[breakable,before skip=1.0mm,after skip=1.0mm,colback=green!3,colframe=green!40!black,title={Tool-Calling Agent},fonttitle=\bfseries,coltitle=white]
{\ttfamily\footnotesize\sloppy Thought:\par\noindent The user wants to exchange a delivered item for another of the same product type with a different option. I need to find the user's user id to proceed with the exchange.\par\noindent \mbox{}\par\noindent Action:\par\noindent \{"name": "find\_\allowbreak{}user\_\allowbreak{}id\_\allowbreak{}by\_\allowbreak{}email",\allowbreak{} "arguments": \{"email": "something@example.com"\}\}}
\end{tcolorbox}

\begin{tcolorbox}[breakable,before skip=1.0mm,after skip=1.0mm,colback=black!2,colframe=black!35,title={Tool Output},fonttitle=\bfseries,coltitle=white]
{\ttfamily\footnotesize\sloppy API output: Error: user not found}
\end{tcolorbox}

\begin{tcolorbox}[breakable,before skip=1.0mm,after skip=1.0mm,colback=green!3,colframe=green!40!black,title={Tool-Calling Agent},fonttitle=\bfseries,coltitle=white]
{\ttfamily\footnotesize\sloppy Thought:\par\noindent The user was not found by email. I need to find the user's user id using their name and zip code.\par\noindent \mbox{}\par\noindent Action:\par\noindent \{"name": "find\_\allowbreak{}user\_\allowbreak{}id\_\allowbreak{}by\_\allowbreak{}name\_\allowbreak{}zip",\allowbreak{} "arguments": \{"first\_\allowbreak{}name": "John",\allowbreak{} "last\_\allowbreak{}name": "Doe",\allowbreak{} "zip": "12345"\}\}}
\end{tcolorbox}

\begin{tcolorbox}[breakable,before skip=1.0mm,after skip=1.0mm,colback=black!2,colframe=black!35,title={Tool Output},fonttitle=\bfseries,coltitle=white]
{\ttfamily\footnotesize\sloppy API output: Error: user not found}
\end{tcolorbox}

\begin{tcolorbox}[breakable,before skip=1.0mm,after skip=1.0mm,colback=green!3,colframe=green!40!black,title={Tool-Calling Agent},fonttitle=\bfseries,coltitle=white]
{\ttfamily\footnotesize\sloppy Thought:\par\noindent The user was not found by name and zip code. I need to ask the user to provide their email or name and zip code to locate their user id.\par\noindent \mbox{}\par\noindent Action:\par\noindent \{"name": "respond",\allowbreak{} "arguments": \{"content": "I'm sorry,\allowbreak{} I couldn't find your account. Could you please provide your email address or your full name and zip code so I can assist you further?"\}\}}
\end{tcolorbox}

\begin{tcolorbox}[breakable,before skip=1.0mm,after skip=1.0mm,colback=blue!3,colframe=blue!45!black,title={User Agent},fonttitle=\bfseries,coltitle=white]
{\ttfamily\footnotesize\sloppy Sure, my name is Sofia Li, and I live in San Antonio, 78260. Can you look up my account with that information?}
\end{tcolorbox}

\begin{tcolorbox}[breakable,before skip=1.0mm,after skip=1.0mm,colback=green!3,colframe=green!40!black,title={Tool-Calling Agent},fonttitle=\bfseries,coltitle=white]
{\ttfamily\footnotesize\sloppy Thought:\par\noindent The user provided their name and zip code. I can use this information to find their user id.\par\noindent \mbox{}\par\noindent Action:\par\noindent \{"name": "find\_\allowbreak{}user\_\allowbreak{}id\_\allowbreak{}by\_\allowbreak{}name\_\allowbreak{}zip",\allowbreak{} "arguments": \{"first\_\allowbreak{}name": "Sofia",\allowbreak{} "last\_\allowbreak{}name": "Li",\allowbreak{} "zip": "78260"\}\}}
\end{tcolorbox}

\begin{tcolorbox}[breakable,before skip=1.0mm,after skip=1.0mm,colback=black!2,colframe=black!35,title={Tool Output},fonttitle=\bfseries,coltitle=white]
{\ttfamily\footnotesize\sloppy API output: sofia\_\allowbreak{}li\_\allowbreak{}9219}
\end{tcolorbox}

\begin{tcolorbox}[breakable,before skip=1.0mm,after skip=1.0mm,colback=green!3,colframe=green!40!black,title={Tool-Calling Agent},fonttitle=\bfseries,coltitle=white]
{\ttfamily\footnotesize\sloppy Thought:\par\noindent I have located the user's account. I can now proceed to assist with the exchange request.\par\noindent \mbox{}\par\noindent Action:\par\noindent \{"name": "get\_\allowbreak{}user\_\allowbreak{}details",\allowbreak{} "arguments": \{"user\_\allowbreak{}id": "sofia\_\allowbreak{}li\_\allowbreak{}9219"\}\}}
\end{tcolorbox}

\begin{tcolorbox}[breakable,before skip=1.0mm,after skip=1.0mm,colback=black!2,colframe=black!35,title={Tool Output},fonttitle=\bfseries,coltitle=white]
{\ttfamily\footnotesize\sloppy API output:\par\noindent \{"payment\_\allowbreak{}methods": \{"paypal\_\allowbreak{}8194385": \{"source": "paypal",\allowbreak{} "id": "paypal\_\allowbreak{}8194385"\},\allowbreak{} "credit\_\allowbreak{}card\_\allowbreak{}3951670": \{"source": "credit\_\allowbreak{}card",\allowbreak{} "brand": "visa",\allowbreak{} "last\_\allowbreak{}four": "6791",\allowbreak{} "id": "credit\_\allowbreak{}card\_\allowbreak{}3951670"\},\allowbreak{} "credit\_\allowbreak{}card\_\allowbreak{}8105988": \{"source": "credit\_\allowbreak{}card",\allowbreak{} "brand": "mastercard",\allowbreak{} "last\_\allowbreak{}four": "8484",\allowbreak{} "id": "credit\_\allowbreak{}card\_\allowbreak{}8105988"\}\},\allowbreak{} "orders": ["\#W4689314",\allowbreak{} "\#W8855135",\allowbreak{} "\#W3916020",\allowbreak{} "\#W5416052"]\}\par\noindent [... name,\allowbreak{} address,\allowbreak{} and email metadata omitted ...]}
\end{tcolorbox}

\begin{tcolorbox}[breakable,before skip=1.0mm,after skip=1.0mm,colback=green!3,colframe=green!40!black,title={Tool-Calling Agent},fonttitle=\bfseries,coltitle=white]
{\ttfamily\footnotesize\sloppy Thought:\par\noindent I have retrieved the user's details,\allowbreak{} including their orders. I need to get the details of the order to proceed with the exchange.\par\noindent \mbox{}\par\noindent Action:\par\noindent \{"name": "get\_\allowbreak{}order\_\allowbreak{}details",\allowbreak{} "arguments": \{"order\_\allowbreak{}id": "\#W4689314"\}\}}
\end{tcolorbox}

\begin{tcolorbox}[breakable,before skip=1.0mm,after skip=1.0mm,colback=black!2,colframe=black!35,title={Tool Output},fonttitle=\bfseries,coltitle=white]
{\ttfamily\footnotesize\sloppy API output:\par\noindent \{"order\_\allowbreak{}id": "\#W4689314",\allowbreak{} "items": [\{"name": "Digital Camera",\allowbreak{} "product\_\allowbreak{}id": "8940227892",\allowbreak{} "item\_\allowbreak{}id": "5996159312",\allowbreak{} "price": 2895.55,\allowbreak{} "options": \{"resolution": "24MP",\allowbreak{} "zoom": "3x",\allowbreak{} "storage": "SD card"\}\}],\allowbreak{} "status": "delivered",\allowbreak{} "payment\_\allowbreak{}history": [\{"transaction\_\allowbreak{}type": "payment",\allowbreak{} "amount": 2895.55,\allowbreak{} "payment\_\allowbreak{}method\_\allowbreak{}id": "credit\_\allowbreak{}card\_\allowbreak{}8105988"\}]\}\par\noindent [... user,\allowbreak{} address,\allowbreak{} and fulfillment metadata omitted ...]}
\end{tcolorbox}

\begin{tcolorbox}[breakable,before skip=1.0mm,after skip=1.0mm,colback=green!3,colframe=green!40!black,title={Tool-Calling Agent},fonttitle=\bfseries,coltitle=white]
{\ttfamily\footnotesize\sloppy Thought:\par\noindent The order is delivered,\allowbreak{} and the user wants to exchange the item for a model with maximum zoom capacity. I need to get the product details to find an item with the maximum zoom capacity.\par\noindent \mbox{}\par\noindent Action:\par\noindent \{"name": "get\_\allowbreak{}product\_\allowbreak{}details",\allowbreak{} "arguments": \{"product\_\allowbreak{}id": "8940227892"\}\}}
\end{tcolorbox}

\begin{tcolorbox}[breakable,before skip=1.0mm,after skip=1.0mm,colback=black!2,colframe=black!35,title={Tool Output},fonttitle=\bfseries,coltitle=white]
{\ttfamily\footnotesize\sloppy API output:\par\noindent \{"name": "Digital Camera",\allowbreak{} "product\_\allowbreak{}id": "8940227892",\allowbreak{} "variants": \{"3892645120": \{"item\_\allowbreak{}id": "3892645120",\allowbreak{} "options": \{"resolution": "30MP",\allowbreak{} "zoom": "10x",\allowbreak{} "storage": "CF card"\},\allowbreak{} "available": false,\allowbreak{} "price": 3070.64\},\allowbreak{} "5996159312": \{"item\_\allowbreak{}id": "5996159312",\allowbreak{} "options": \{"resolution": "24MP",\allowbreak{} "zoom": "3x",\allowbreak{} "storage": "SD card"\},\allowbreak{} "available": true,\allowbreak{} "price": 2895.55\},\allowbreak{} "7583936705": \{"item\_\allowbreak{}id": "7583936705",\allowbreak{} "options": \{"resolution": "20MP",\allowbreak{} "zoom": "10x",\allowbreak{} "storage": "CF card"\},\allowbreak{} "available": false,\allowbreak{} "price": 3101.43\},\allowbreak{} "5484530610": \{"item\_\allowbreak{}id": "5484530610",\allowbreak{} "options": \{"resolution": "24MP",\allowbreak{} "zoom": "10x",\allowbreak{} "storage": "CF card"\},\allowbreak{} "available": false,\allowbreak{} "price": 3109.83\},\allowbreak{} "9228757377": \{"item\_\allowbreak{}id": "9228757377",\allowbreak{} "options": \{"resolution": "30MP",\allowbreak{} "zoom": "10x",\allowbreak{} "storage": "SD card"\},\allowbreak{} "available": true,\allowbreak{} "price": 3066.23\}\}\}\par\noindent [... 9 other 3x/\allowbreak{}5x camera variants omitted ...]}
\end{tcolorbox}

\begin{tcolorbox}[breakable,before skip=1.0mm,after skip=1.0mm,colback=green!3,colframe=green!40!black,title={Tool-Calling Agent},fonttitle=\bfseries,coltitle=white]
{\ttfamily\footnotesize\sloppy Thought:\par\noindent I have found an available item with a 10x zoom capacity. I need to confirm the exchange details with the user before proceeding.\par\noindent \mbox{}\par\noindent Action:\par\noindent \{"name": "respond",\allowbreak{} "arguments": \{"content": "I found an available digital camera with a 10x zoom capacity. The item ID is 9228757377. The price is \$3066.23. The price difference will be \$170.68. Would you like to proceed with this exchange?"\}\}}
\end{tcolorbox}

\begin{tcolorbox}[breakable,before skip=1.0mm,after skip=1.0mm,colback=blue!3,colframe=blue!45!black,title={User Agent},fonttitle=\bfseries,coltitle=white]
{\ttfamily\footnotesize\sloppy That’s not quite what I was looking for. I need a camera with the maximum zoom capacity possible, ideally something around 60x or higher. Could you check if there’s a model like that available? Price isn’t an issue, but I want to keep all other specifications the same as my current camera.}
\end{tcolorbox}

\begin{tcolorbox}[breakable,before skip=1.0mm,after skip=1.0mm,colback=green!3,colframe=green!40!black,title={Tool-Calling Agent},fonttitle=\bfseries,coltitle=white]
{\ttfamily\footnotesize\sloppy Thought:\par\noindent The user is looking for a camera with a maximum zoom capacity of 60x or higher while keeping other specifications the same. I need to check the product details again to find such an item.\par\noindent \mbox{}\par\noindent Action:\par\noindent \{"name": "get\_\allowbreak{}product\_\allowbreak{}details",\allowbreak{} "arguments": \{"product\_\allowbreak{}id": "8940227892"\}\}}
\end{tcolorbox}

\begin{tcolorbox}[breakable,before skip=1.0mm,after skip=1.0mm,colback=black!2,colframe=black!35,title={Tool Output},fonttitle=\bfseries,coltitle=white]
{\ttfamily\footnotesize\sloppy API output:\par\noindent \{"name": "Digital Camera",\allowbreak{} "product\_\allowbreak{}id": "8940227892"\}\par\noindent [... repeated inventory output omitted; unchanged from the previous tool output ...]}
\end{tcolorbox}

\begin{tcolorbox}[breakable,before skip=1.0mm,after skip=1.0mm,colback=green!3,colframe=green!40!black,title={Tool-Calling Agent},fonttitle=\bfseries,coltitle=white]
{\ttfamily\footnotesize\sloppy Thought:\par\noindent After reviewing the available options,\allowbreak{} there is no item with a zoom capacity of 60x or higher. The highest available zoom capacity is 10x. I will inform the user of this and ask if they would like to proceed with the 10x zoom option.\par\noindent \mbox{}\par\noindent Action:\par\noindent \{"name": "respond",\allowbreak{} "arguments": \{"content": "I'm sorry,\allowbreak{} but there is no available digital camera with a zoom capacity of 60x or higher. The highest available zoom capacity is 10x. Would you like to proceed with the 10x zoom option?"\}\}}
\end{tcolorbox}

\begin{tcolorbox}[breakable,before skip=1.0mm,after skip=1.0mm,colback=blue!3,colframe=blue!45!black,title={User Agent},fonttitle=\bfseries,coltitle=white]
{\ttfamily\footnotesize\sloppy I see. Well, if 10x is the best you can offer, then I’ll have to consider other options. However, before I decide, could you confirm if there are any other models with a higher zoom capacity, even if they’re not exactly the same as my current camera? I really need the extra zoom for bird-watching.}
\end{tcolorbox}

\begin{tcolorbox}[breakable,before skip=1.0mm,after skip=1.0mm,colback=green!3,colframe=green!40!black,title={Tool-Calling Agent},fonttitle=\bfseries,coltitle=white]
{\ttfamily\footnotesize\sloppy Thought:\par\noindent The user is looking for a camera with a higher zoom capacity than the available 10x option. I need to check if there are any other models with a higher zoom capacity,\allowbreak{} even if they are not the same as the current camera.\par\noindent \mbox{}\par\noindent Action:\par\noindent \{"name": "get\_\allowbreak{}product\_\allowbreak{}details",\allowbreak{} "arguments": \{"product\_\allowbreak{}id": "8940227892"\}\}}
\end{tcolorbox}

\begin{tcolorbox}[breakable,before skip=1.0mm,after skip=1.0mm,colback=black!2,colframe=black!35,title={Tool Output},fonttitle=\bfseries,coltitle=white]
{\ttfamily\footnotesize\sloppy API output:\par\noindent \{"name": "Digital Camera",\allowbreak{} "product\_\allowbreak{}id": "8940227892"\}\par\noindent [... repeated inventory output omitted; unchanged from the previous tool output ...]}
\end{tcolorbox}

\begin{tcolorbox}[breakable,before skip=1.0mm,after skip=1.0mm,colback=green!3,colframe=green!40!black,title={Tool-Calling Agent},fonttitle=\bfseries,coltitle=white]
{\ttfamily\footnotesize\sloppy Thought:\par\noindent After reviewing the available options,\allowbreak{} there are no models with a zoom capacity higher than 10x. I will inform the user of this and ask if they would like to proceed with the 10x zoom option or if they need further assistance.\par\noindent \mbox{}\par\noindent Action:\par\noindent \{"name": "respond",\allowbreak{} "arguments": \{"content": "I'm sorry,\allowbreak{} but there are no models with a zoom capacity higher than 10x available. The highest available zoom capacity is 10x. Would you like to proceed with the 10x zoom option,\allowbreak{} or is there anything else I can assist you with?"\}\}}
\end{tcolorbox}

\begin{tcolorbox}[breakable,before skip=1.0mm,after skip=1.0mm,colback=blue!3,colframe=blue!45!black,title={User Agent},fonttitle=\bfseries,coltitle=white]
{\ttfamily\footnotesize\sloppy Alright, thank you for checking. Since 10x is the best option available, I’ll go ahead with the exchange. Can we process this quickly, and I’ll use my PayPal account to cover the additional cost.}
\end{tcolorbox}

\begin{tcolorbox}[breakable,before skip=1.0mm,after skip=1.0mm,colback=green!3,colframe=green!40!black,title={Tool-Calling Agent},fonttitle=\bfseries,coltitle=white]
{\ttfamily\footnotesize\sloppy Thought:\par\noindent The user has confirmed the exchange and provided the payment method. I can now proceed with the exchange of the delivered order item.\par\noindent \mbox{}\par\noindent Action:\par\noindent \{"name": "exchange\_\allowbreak{}delivered\_\allowbreak{}order\_\allowbreak{}items",\allowbreak{} "arguments": \{"order\_\allowbreak{}id": "\#W4689314",\allowbreak{} "item\_\allowbreak{}ids": ["5996159312"],\allowbreak{} "new\_\allowbreak{}item\_\allowbreak{}ids": ["9228757377"],\allowbreak{} "payment\_\allowbreak{}method\_\allowbreak{}id": "paypal\_\allowbreak{}8194385"\}\}}
\end{tcolorbox}

\begin{tcolorbox}[breakable,before skip=1.0mm,after skip=1.0mm,colback=black!2,colframe=black!35,title={Tool Output},fonttitle=\bfseries,coltitle=white]
{\ttfamily\footnotesize\sloppy API output:\par\noindent \{"order\_\allowbreak{}id": "\#W4689314",\allowbreak{} "status": "exchange requested",\allowbreak{} "exchange\_\allowbreak{}items": ["5996159312"],\allowbreak{} "exchange\_\allowbreak{}new\_\allowbreak{}items": ["9228757377"],\allowbreak{} "exchange\_\allowbreak{}payment\_\allowbreak{}method\_\allowbreak{}id": "paypal\_\allowbreak{}8194385",\allowbreak{} "exchange\_\allowbreak{}price\_\allowbreak{}difference": 170.68\}\par\noindent [... unchanged user,\allowbreak{} address,\allowbreak{} item,\allowbreak{} fulfillment,\allowbreak{} and payment-history fields omitted ...]}
\end{tcolorbox}

\begin{tcolorbox}[breakable,before skip=1.0mm,after skip=1.0mm,colback=green!3,colframe=green!40!black,title={Tool-Calling Agent},fonttitle=\bfseries,coltitle=white]
{\ttfamily\footnotesize\sloppy Thought:\par\noindent The exchange has been successfully processed. I will inform the user about the next steps and confirm the details.\par\noindent \mbox{}\par\noindent Action:\par\noindent \{"name": "respond",\allowbreak{} "arguments": \{"content": "The exchange has been processed successfully. The item with ID 5996159312 will be exchanged for the item with ID 9228757377. The price difference of \$170.68 will be covered by your PayPal account. You will receive an email with further instructions on how to return the item. Is there anything else I can assist you with?"\}\}}
\end{tcolorbox}

\begin{tcolorbox}[breakable,before skip=1.0mm,after skip=1.0mm,colback=blue!3,colframe=blue!45!black,title={User Agent},fonttitle=\bfseries,coltitle=white]
{\ttfamily\footnotesize\sloppy Great, thank you for your help! I appreciate the quick processing. That’s all I need for now. Have a great day! \#\#\#STOP\#\#\#}
\end{tcolorbox}

\begin{tcolorbox}[breakable,before skip=1.0mm,after skip=1.0mm,colback=bestyellow!65,colframe=orange!55!black,title={Task outcome},fonttitle=\bfseries,coltitle=white]
{\ttfamily\footnotesize\sloppy Reward = 1. CATC-8B succeeds in all four trials of the reported run (4/4).}
\end{tcolorbox}

\subsection{CATC-8B+CAST-8B Task Trajectory}
\label{app:task52_cast_intervention}

\begin{tcolorbox}[breakable,before skip=1.0mm,after skip=1.0mm,colback=blue!3,colframe=blue!45!black,title={User Agent},fonttitle=\bfseries,coltitle=white]
{\ttfamily\footnotesize\sloppy That sounds good, but I was hoping for a camera with even more zoom capacity. Is there a model with a higher zoom level available? Price isn’t an issue for me.}
\end{tcolorbox}

\begin{tcolorbox}[breakable,before skip=1.0mm,after skip=1.0mm,colback=orange!4,colframe=orange!60!black,title={Tool-Calling Agent: Proposed Action},fonttitle=\bfseries,coltitle=white]
{\ttfamily\footnotesize\sloppy Thought:\par\noindent The user is looking for a digital camera with the maximum zoom capacity available. I have already identified the highest available zoom capacity (10x) for the digital camera. Since the user mentioned that price is not an issue,\allowbreak{} I can proceed with the exchange for the 10x zoom model.\par\noindent \mbox{}\par\noindent Action:\par\noindent \{"name": "exchange\_\allowbreak{}delivered\_\allowbreak{}order\_\allowbreak{}items",\allowbreak{} "arguments": \{"order\_\allowbreak{}id": "\#W4689314",\allowbreak{} "item\_\allowbreak{}ids": ["5996159312"],\allowbreak{} "new\_\allowbreak{}item\_\allowbreak{}ids": ["9228757377"],\allowbreak{} "payment\_\allowbreak{}method\_\allowbreak{}id": "credit\_\allowbreak{}card\_\allowbreak{}8105988"\}\}}
\end{tcolorbox}

\begin{tcolorbox}[breakable,before skip=1.0mm,after skip=1.0mm,colback=red!3,colframe=red!45!black,title={Critque Agent: Rationale},fonttitle=\bfseries,coltitle=white]
{\ttfamily\footnotesize\sloppy Let's verify step by step. The user explicitly requested a camera with 'even more zoom capacity' than the 10x model already identified as the maximum available in the product details (which showed 10x as the highest available). The assistant's thought process claims it identified the 'highest available zoom capacity (10x)', but the user is asking for something *more* than that. The assistant then proceeds to call the exchange tool for the 10x model. This violates the policy requiring the agent to 'explain the exchange detail and ask for explicit user confirmation (yes/no) to proceed' before taking action. The assistant failed to explain that no higher zoom option exists and did not obtain explicit confirmation for the specific item (10x) before executing the tool. Additionally, the assistant's action is incorrect because it executes the tool immediately without this required confirmation step, and the item selected (10x) is not the one the user requested (which would be impossible given the data, but the assistant should have clarified the impossibility and confirmed the 10x option as the only available choice). The tool selection is also incorrect for this turn because the correct action should have been to respond to the user explaining the situation and asking for confirmation,}
\end{tcolorbox}

\begin{tcolorbox}[breakable,before skip=1.0mm,after skip=1.0mm,colback=violet!4,colframe=violet!55!black,title={critique Agent: Verification},fonttitle=\bfseries,coltitle=white]
{\ttfamily\footnotesize\sloppy Verification: Is the action correct (Yes/\allowbreak{}No)? No}
\end{tcolorbox}

\begin{tcolorbox}[breakable,before skip=1.0mm,after skip=1.0mm,colback=green!3,colframe=green!45!black,title={Tool-Calling Agent: Revised Action},fonttitle=\bfseries,coltitle=white]
{\ttfamily\footnotesize\sloppy Thought:\par\noindent The user is looking for a digital camera with the maximum zoom capacity available,\allowbreak{} and I have identified that the highest available zoom capacity is 10x. Since no higher zoom option is available,\allowbreak{} I will explain this to the user and ask for their confirmation to proceed with the exchange for the 10x zoom model.\par\noindent \mbox{}\par\noindent Action:\par\noindent \{"name": "respond",\allowbreak{} "arguments": \{"content": "I apologize for the confusion. After reviewing the available options,\allowbreak{} the highest zoom capacity available for the digital camera is 10x. There are no models with a higher zoom capacity than this. The price for the 10x zoom model is \$3066.23,\allowbreak{} which is \$170.68 more than the current item in your order. Would you like to proceed with the exchange for the 10x zoom model?"\}\}}
\end{tcolorbox}

\begin{tcolorbox}[breakable,before skip=1.0mm,after skip=1.0mm,colback=violet!4,colframe=violet!55!black,title={critique Agent: Verification},fonttitle=\bfseries,coltitle=white]
{\ttfamily\footnotesize\sloppy Verification: Is the action correct (Yes/\allowbreak{}No)? Yes}
\end{tcolorbox}

\begin{tcolorbox}[breakable,before skip=1.0mm,after skip=1.0mm,colback=blue!3,colframe=blue!45!black,title={User Agent},fonttitle=\bfseries,coltitle=white]
{\ttfamily\footnotesize\sloppy I understand. In that case, I would like to proceed with the exchange for the 10x zoom model. Can we complete this as soon as possible? I’ll use my PayPal account for the additional payment.}
\end{tcolorbox}

\begin{tcolorbox}[breakable,before skip=1.0mm,after skip=1.0mm,colback=oursblue!55,colframe=blue!45!black,title={Tool-Calling Agent: Action Deployed},fonttitle=\bfseries,coltitle=white]
{\ttfamily\footnotesize\sloppy Thought:\par\noindent The user has confirmed the exchange for the 10x zoom model and provided the payment method. I can now proceed with the exchange.\par\noindent \mbox{}\par\noindent Action:\par\noindent \{"name": "exchange\_\allowbreak{}delivered\_\allowbreak{}order\_\allowbreak{}items",\allowbreak{} "arguments": \{"order\_\allowbreak{}id": "\#W4689314",\allowbreak{} "item\_\allowbreak{}ids": ["5996159312"],\allowbreak{} "new\_\allowbreak{}item\_\allowbreak{}ids": ["9228757377"],\allowbreak{} "payment\_\allowbreak{}method\_\allowbreak{}id": "paypal\_\allowbreak{}8194385"\}\}}
\end{tcolorbox}

\begin{tcolorbox}[breakable,before skip=1.0mm,after skip=1.0mm,colback=violet!4,colframe=violet!55!black,title={Critique Agent: Verification},fonttitle=\bfseries,coltitle=white]
{\ttfamily\footnotesize\sloppy Verification: Is the action correct (Yes/\allowbreak{}No)? Yes}
\end{tcolorbox}

\begin{tcolorbox}[breakable,before skip=1.0mm,after skip=1.0mm,colback=bestyellow!65,colframe=orange!55!black,title={Task outcome},fonttitle=\bfseries,coltitle=white]
{\ttfamily\footnotesize\sloppy Reward = 1 in this representative critique-guided trial. The order reaches 'exchange requested' with item 9228757377 and payment method paypal\_8194385.}
\end{tcolorbox}

\subsection{Verification Failure: Incorrect Verification }
\label{app:incorrect_verification}
We also present a verification failure case for a better assessment of the CAST system. 
This test example came from Retail Task 23 exposes 
The pending order contains grill item \texttt{5946177616}, while item
\texttt{7082455361} belongs to the delivered order whose configuration the
user wants to match. 
The Tool-Calling Agent wrongly reverses the current and target item IDs. 
The critique Agent incorrectly verifies the proposed tool call and missed the reversed IDs, so the action is deployed and failed.

\begin{tcolorbox}[breakable,before skip=1.0mm,after skip=1.0mm,colback=black!2,colframe=black!45,title={Context},fonttitle=\bfseries,coltitle=white]
{\ttfamily\footnotesize\sloppy Pending order \#W3561391 contains grill item 5946177616 (gas, portable, no extra features). Delivered order \#W6876713 contains grill item 7082455361 (charcoal, medium, rotisserie), whose configuration the user wants to match. [... authentication and unrelated exchanges omitted ...]}
\end{tcolorbox}

\begin{tcolorbox}[breakable,before skip=1.0mm,after skip=1.0mm,colback=black!2,colframe=black!35,title={Tool Output},fonttitle=\bfseries,coltitle=white]
{\ttfamily\footnotesize\sloppy API output:\par\noindent \{"name": "Grill",\allowbreak{} "product\_\allowbreak{}id": "6819683148",\allowbreak{} "variants": \{"7082455361": \{"item\_\allowbreak{}id": "7082455361",\allowbreak{} "options": \{"type": "charcoal",\allowbreak{} "size": "medium",\allowbreak{} "features": "rotisserie"\},\allowbreak{} "available": true,\allowbreak{} "price": 962.69\},\allowbreak{} "5946177616": \{"item\_\allowbreak{}id": "5946177616",\allowbreak{} "options": \{"type": "gas",\allowbreak{} "size": "portable",\allowbreak{} "features": "none"\},\allowbreak{} "available": true,\allowbreak{} "price": 1057.24\}\}\}\par\noindent [... 10 unrelated variants omitted ...]}
\end{tcolorbox}

\begin{tcolorbox}[breakable,before skip=1.0mm,after skip=1.0mm,colback=orange!4,colframe=orange!60!black,title={Tool-Calling Agent: Proposed Action},fonttitle=\bfseries,coltitle=white]
{\ttfamily\footnotesize\sloppy Thought:\par\noindent I have identified the correct new item ID for the grill modification to match the one the user already received. I will now proceed with the modification.\par\noindent \mbox{}\par\noindent Action:\par\noindent \{"name": "modify\_\allowbreak{}pending\_\allowbreak{}order\_\allowbreak{}items",\allowbreak{} "arguments": \{"order\_\allowbreak{}id": "\#W3561391",\allowbreak{} "item\_\allowbreak{}ids": ["7082455361"],\allowbreak{} "new\_\allowbreak{}item\_\allowbreak{}ids": ["5946177616"],\allowbreak{} "payment\_\allowbreak{}method\_\allowbreak{}id": "credit\_\allowbreak{}card\_\allowbreak{}7901829"\}\}}
\end{tcolorbox}

\begin{tcolorbox}[breakable,before skip=1.0mm,after skip=1.0mm,colback=red!3,colframe=red!45!black,title={Critique Agent: Rationale},fonttitle=\bfseries,coltitle=white]
{\ttfamily\footnotesize\sloppy Let's verify step by step. The user wants to modify the grill in order \#W3561391 to match the one they already received. The assistant correctly identified the current item (7082455361) and the target item (5946177616) based on the product details retrieved in the previous turn. The tool selected, `modify\_pending\_order\_items`, is the correct tool for this action. The arguments provided (order\_id, item\_ids, new\_item\_ids, payment\_method\_id) are accurate and grounded in the conversation history and tool outputs. There are no domain policy violations, as the order is pending and the item is being modified to a variant of the same product. No hallucinations are present. The action is correct.}
\end{tcolorbox}

\begin{tcolorbox}[breakable,before skip=1.0mm,after skip=1.0mm,colback=violet!4,colframe=violet!55!black,title={Critique Agent: Verification},fonttitle=\bfseries,coltitle=white]
{\ttfamily\footnotesize\sloppy Verification: Is the action correct (Yes/\allowbreak{}No)? Yes}
\end{tcolorbox}

\begin{tcolorbox}[breakable,before skip=1.0mm,after skip=1.0mm,colback=oursblue!55,colframe=blue!45!black,title={Tool-Calling Agent: Action Deployed},fonttitle=\bfseries,coltitle=white]
{\ttfamily\footnotesize\sloppy Thought:\par\noindent I have identified the correct new item ID for the grill modification to match the one the user already received. I will now proceed with the modification.\par\noindent \mbox{}\par\noindent Action:\par\noindent \{"name": "modify\_\allowbreak{}pending\_\allowbreak{}order\_\allowbreak{}items",\allowbreak{} "arguments": \{"order\_\allowbreak{}id": "\#W3561391",\allowbreak{} "item\_\allowbreak{}ids": ["7082455361"],\allowbreak{} "new\_\allowbreak{}item\_\allowbreak{}ids": ["5946177616"],\allowbreak{} "payment\_\allowbreak{}method\_\allowbreak{}id": "credit\_\allowbreak{}card\_\allowbreak{}7901829"\}\}}
\end{tcolorbox}

\begin{tcolorbox}[breakable,before skip=1.0mm,after skip=1.0mm,colback=black!2,colframe=black!35,title={Tool Output},fonttitle=\bfseries,coltitle=white]
{\ttfamily\footnotesize\sloppy Error: 7082455361 not found}
\end{tcolorbox}

\begin{tcolorbox}[breakable,before skip=1.0mm,after skip=1.0mm,colback=bestyellow!65,colframe=orange!55!black,title={Task outcome},fonttitle=\bfseries,coltitle=white]
{\ttfamily\footnotesize\sloppy Reward = 0. The approved call fails because item 7082455361 is not present in pending order \#W3561391.}
\end{tcolorbox}

\section{License}
This work uses models and benchmarks from various sources. The following table lists the licenses for all models and benchmarks used in this study:

\begin{table}[h]
\centering
\begin{tabular}{ll}
\toprule
\textbf{Model/Benchmark} & \textbf{License Type} \\
\midrule
Qwen3-4B & Apache 2.0 \\
Qwen3-8B & Apache 2.0 \\
Qwen3-32B & Apache 2.0 \\
Qwen3-235B-A22B & Apache 2.0 \\
Qwen3.5-27B & Apache 2.0 \\
Qwen3.5-122B-A10B & Apache 2.0 \\
Qwen2.5-72B & Qwen License \\
GPT-4.1 & Proprietary (OpenAI) \\
GPT-OSS-120B & Apache 2.0 \\
$\tau$-Bench & MIT \\
$\tau$-Trait & MIT \\
\bottomrule
\end{tabular}
\end{table}

% \section{Examples}

\end{document}